\documentclass[twocolumn]{svjour3}          % twocolumn
\usepackage{graphicx}

\usepackage{amsmath}
\usepackage{amssymb}
\usepackage{cite}

\usepackage{array}
\usepackage{stfloats}
\usepackage{url}
\usepackage{natbib}

\usepackage{nicefrac}
\usepackage{mathptmx}      % use Times fonts if available on your TeX system
\usepackage{latexsym}
\usepackage{hyperref}
\usepackage{breakurl}
\usepackage{breakcites}
\usepackage{rotating}
\usepackage{multirow}
\usepackage{xspace}
\usepackage{boldline}
\usepackage{makecell}
\usepackage{color}
\usepackage{colortbl}
\def\eg{\emph{e.g.}\xspace}
\def\ie{\emph{i.e.}\xspace}

\def\etal{\emph{et al.}\xspace}

\begin{document}

\title{Visiting the Invisible: Layer-by-Layer Completed Scene Decomposition%\thanks{This research was supported by the BeingTogether Centre, a collaboration between Nanyang Technological University (NTU) Singapore and University of North Carolina (UNC) at Chapel Hill. The BeingTogether Centre was supported by the National Research Foundation, Prime Minister’s Office, Singapore under its International Research Centres in Singapore Funding Initiative. This research was also conducted in collaboration with Singapore Telecommunications Limited and partially supported by the Singapore Government through the Industry Alignment Fund ‐- Industry Collaboration Projects Grant.}
}


\author{Chuanxia~Zheng \and
	    Duy-Son Dao \and
	    Guoxian Song \and
        Tat-Jen~Cham \and
        Jianfei~Cai
}


\institute{ Chuanxia~Zheng \and Guoxian Song \and Tat-Jen~Cham\at
              School of Computer Science and Engineering, Nanyang Technological University, Singapore. \\
              \email{chuanxia001@e.ntu.edu.sg, guoxian001@e.ntu.edu.sg astjcham@ntu.edu.sg}           %  \\
           \and
           Duy-Son Dao \and Jianfei~Cai \at
           Department of Data Science \& AI, Monash University, Australia.
           \email{duy.dao@monash.edu, jianfei.cai@monash.edu}
}


\maketitle
\begin{abstract}
	
Existing scene understanding systems mainly focus on recognizing the visible parts of a scene, ignoring the intact appearance of physical objects in the real-world. Concurrently, image completion has aimed to create plausible appearance for the invisible regions, but requires a manual mask as input. In this work, we propose a higher-level scene understanding system to tackle both visible and invisible parts of objects and backgrounds in a given scene. Particularly, we built a system to decompose a scene into individual objects, infer their underlying occlusion relationships, and even automatically learn which parts of the objects are occluded that need to be completed. In order to disentangle the occluded relationships of all objects in a complex scene, we use the fact that the front object without being occluded is easy to be identified, detected, and segmented. Our system interleaves the two tasks of instance segmentation and scene completion through multiple iterations, solving for objects layer-by-layer. We first provide a thorough experiment using a new realistically rendered dataset with ground-truths for all invisible regions. To bridge the domain gap to real imagery where ground-truths are unavailable, we then train another model with the pseudo-ground-truths generated from our trained synthesis model. We demonstrate results on a wide variety of datasets and show significant improvement over the state-of-the-art. The code will be available at \href{https://github.com/lyndonzheng/VINV}{https://github.com/lyndonzheng/VINV}.

\end{abstract}

\begin{figure*}[tb!]
	\centering
	\includegraphics[width=\linewidth,height=0.18\textheight]{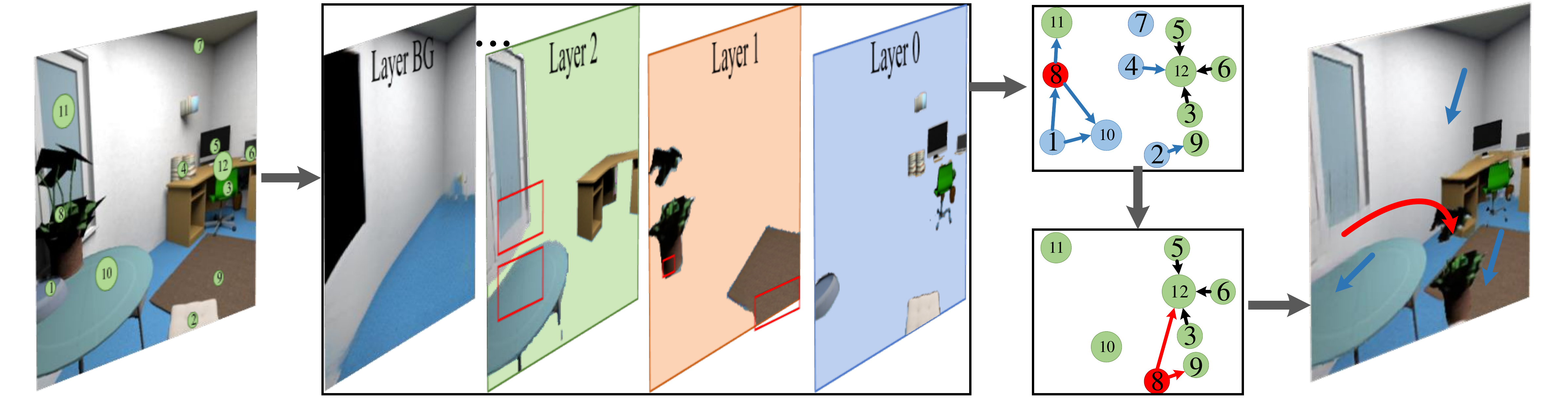}
	\begin{picture}(0,0)
		\put(-210,0){(a) Input}
		\put(-120,0){(b) Layered Completed Scene Decomposition}
		\put(80, 0){(c) Pairwise Order}
		\put(155, 0){(d) Image Recomposition}
	\end{picture}
	\caption{\textbf{Example results of scene decomposition and recomposition.} (a) Input. (b) Our model structurally decomposes a scene into individual completed objects. Red rectangles highlight the original \emph{invisible} parts. (c) The inferred pairwise order (top graph) and edited order (bottom graph) of the instances. Blue nodes indicate the deleted objects while the red node is the moved object. (d) The new recomposed scene.}
	\label{fig:example}
\end{figure*}
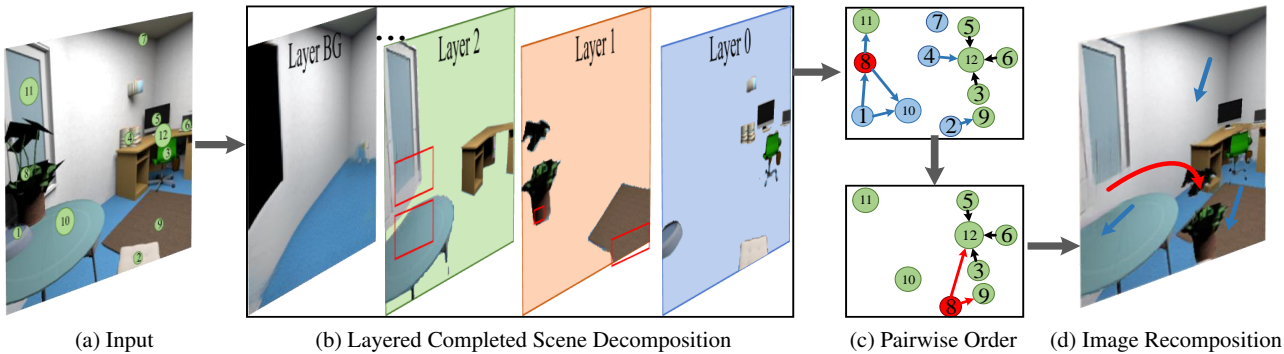

\section{Introduction}\label{sec:intro}

The vision community has made rapid advances in scene understanding tasks, such as object classification and localization~\citep{girshick2014rich,he2015spatial,ren2015faster}, scene parsing~\citep{long2015fully,chen2017deeplab,badrinarayanan2017segnet}, instance segmentation~\citep{pinheiro2015learning,he2017mask,chen2019hybrid}, and layered scene decomposition~\citep{gould2009decomposing,yang2010layered,zhang2015monocular}. Despite their impressive performance, these systems deal only with \emph{visible} parts of scenes without trying to exploit \emph{invisible} regions, which results in an uncompleted representation of real objects. 

In parallel, significantly progress for the generation task has been made with the emergence of deep generative networks, such as GAN-based models \citep{goodfellow2014generative,gulrajani2017improved,karras2019style}, VAE-based models \citep{kingma2013auto,van2017neural,vahdat2020NVAE}, and flow-based models \citep{dinh2014nice,dinh2016density,kingma2018glow}. Empowered by these techniques, image completion~\citep{,iizuka2017globally,yu2018generative,zheng2019pluralistic} and object completion~\citep{ehsani2018segan,zhan2020self,ling2020variational} have made it possible to create the plausible appearances for occluded objects and backgrounds. However, these systems depend on manual masks or visible ground-truth masks as input, rather than automatically understand the full scene.

In this paper, we aim to build a system that has the ability to \emph{decompose} a scene into individual objects, \emph{infer} their underlying occlusion relationships, and moreover \emph{imagine} what occluded objects may look like, \emph{while using only an image as input}. This novel task involves the classical recognition task of instance segmentation to predict the geometry and category of all objects in a scene, and the generation task of image completion to reconstruct invisible parts of objects and backgrounds.

To decompose a scene into instances with completed appearances in one pass is extremely challenging. This is because realistic natural scenes often consist of a vast collection of physical objects, with complex scene structure and occlusion relationships, especially when one object is occluded by multiple objects, or when instances have deep hierarchical occlusion relationships. 

Our core idea is from the observation that \emph{it is much easier to identify, detect and segment foreground objects than occluded objects}. Motivated by this, we propose a {\bf C}ompleted {\bf S}cene {\bf D}ecomposition {\bf Net}work ({\bf CSDNet}) that learns to segment and complete each object in a scene layer-by-layer consecutively. As shown in Fig.~\ref{fig:example}, our layered scene decomposition network only segments the fully visible objects out in each layer (Fig.~\ref{fig:example}(b)). If the system is able to properly segment the foreground objects, it will automatically learn which parts of occluded objects are actually invisible that need to be filled in. The completed image is then passed back to the layered scene decomposition network, which can again focus purely on detecting and segmenting visible objects. As the interleaving proceeds, a structured instance depth order (Fig.~\ref{fig:example}(c)) is progressively derived by using the inferred absolute layer order. The thorough decomposition of a scene along with spatial relationships allows the system to freely recompose a new scene (Fig.~\ref{fig:example}(d)).

Another challenge in this novel task is the lack of data: there is no complex, realistic dataset that provides intact ground-truth appearance for originally occluded objects and backgrounds in a scene. While latest works~\citep{li2016amodal,zhan2020self} introduced a self-supervised way to tackle the amodal completion using only visible annotations, they can not do a fair quantitative comparison as no real ground-truths are available. To mitigate this issue, we constructed a high-quality rendered dataset, named \textbf{C}ompleted \textbf{S}cene \textbf{D}ecomposition (\textbf{CSD}), based on more than 2k indoor rooms. Unlike the datasets in \citep{ehsani2018segan,Dhamo2019iccv}, our dataset is designed to have more typical camera viewpoints, with near-realistic appearance.

As elaborated in Section~\ref{sec:csd_results}, the proposed system performs well on this rendered dataset, both qualitatively and quantitatively outperforming existing methods in completed scene decomposition, in terms of instance segmentation, depth ordering, and amodal mask and content completion. To further demonstrate the generalization of our system, we extend it to real datasets. As there is no ground truth annotations and appearance available for training, we created pseudo-ground-truths for real images using our model that is purely trained on \textbf{CSD}, and then fine-tuned this model accordingly. This model outperforms state-of-the-art methods~\citep{zhu2017semantic,qi2019amodal,zhan2020self} on amodal instance segmentation and depth ordering tasks, despite these methods being specialized to their respective tasks rather than our holistic completed scene decomposition task. %While we are unable to quantitatively evaluate real-image scene completion without ground truth appearance for occluded objects, our method is able to create visually reasonable layer-by-layer decomposition results, and we further demonstrate its effectiveness in real scene recomposition. 

In summary, we propose a layer-by-layer scene decomposition network that jointly learns structural scene decomposition and completion, rather than treating them separately as the existing works \citep{ehsani2018segan,Dhamo2019iccv,zhan2020self}. To our knowledge, it is the first work that proposes to complete objects based on the global context, instead of tackling each object independently. To address this novel task, we render a high-quality rendered dataset with ground-truth for all instances. We then provide a thorough ablation study using this rendered dataset, in which we demonstrate that the method substantially outperforms existing methods that address the task in isolation. On real images, we improve the performance to the recent state-of-the-art methods by using pseudo-ground-truth as weakly-supervised labels. The experimental results show that our \textbf{CSDNet} is able to acquire a full decomposition of a scene, \emph{with only an image as input}, which conduces to a lot of applications, \eg object-level image editing.

The rest of the paper is organized as follows. We discuss the related work in Section~\ref{sec:related}, and describe our layer-by-layer CSDNet in detail in Section~\ref{sec:method}. In Section~\ref{sec:data} we present our rendered dataset. We then show the experiment results on this synthetic dataset as well as the results on real-world images in Section~\ref{sec:experiment}, followed by a conclusion in Section~\ref{sec:conclusions}.

\begin{table}[tb!]
	\caption{Comparison with related work based on three aspects: outputs, inputs and data. I: image, In: inmodal segmentation, O: occlusion order, SP: scene parsing, AB: amodal bounding box, AS: amodal surface, A: amodal segmentation, D: depth, IRGB: intact RGB object.}
	\label{tab:related_work}
	\scriptsize
	\renewcommand{\arraystretch}{1.3}
	\begin{tabular}{@{}llll@{}}
		\hlineB{3.5}
		Paper & Outputs & Inputs & Data\\
		\hline
		 & SP, O & I & LabelMe, PASVOC, others \\
		\citep{yang2011layered} & In, O & I & PASVOC \\
		\citep{tighe2014scene} & SP, O & I & LabelMe, SUN \\
		\citep{zhang2015monocular} & In, O & I & KITTI \\
		\hline
		\citep{guo2012beyond} & AS & I & StreetScenes, SUN, others\\
		\citep{kar2015amodal} & AB & I & PASVOC, PAS3D \\
		\citep{liu2016layered} & AS, O & I, D & NTUv2-D\\
		\hline
		\citep{li2016amodal} & A & I, In & PASVOC\\
		\citep{zhu2017semantic} & A, O & I & COCOA (from COCO)\\
		\citep{follmann2019learning} & A & I & COCOA, COCOA-cls, D2S\\
		\citep{qi2019amodal} & A & I & KINS (from KITTI) \\
		\citep{hu2019sail} & A & I & Synthesis video\\
		\hline
		\citep{ehsani2018segan} & A, O, IRGB & I, In & DYCE, PAS3D\\
		\citep{zhan2020self} & A, O, IRGB & I, In & KINS, COCOA \\
		\citep{ling2020variational} & A, IRGB & I, In & KINS\\
		\citep{yan2019visualizing} & A, IRGB & I & Vehicle \\
		\citep{burgess2019monet} & In, IRGB & I &  Toy\\
		\citep{Dhamo2019iccv} & A, D, IRGB& I & SUNCG, Stanford 2D-3D\\
		Ours & A, O, IRGB& I & KINS, COCOA, SUNCG\\
		\hlineB{2.5}
	\end{tabular}
\end{table}

\section{Related Work}\label{sec:related}
A variety of scene understanding tasks have previously been proposed, including layered scene decomposition~\citep{yang2011layered}, instance segmentation~\citep{he2017mask}, amodal segmentation~\citep{li2016amodal}, and scene parsing~\citep{chen2017deeplab}. In order to clarify the relationships of our work to the relevant literature, Table~\ref{tab:related_work} gives a comparison based on three aspects: what the goals are, which information is used, and on which dataset is evaluated.

\par\medskip\noindent\textbf{Layered scene decomposition for inmodal perception.} The layered scene decomposition for visible regions has been extensively studied in the literature. Shade \etal \citep{shade1998layered} first proposed a representation called a layered depth image (LDI), which contains multiple layers for a complex scene. Based on this image representation that requires occlusion reasoning, the early works focused on ordering the semantic map as occluded and visible regions. Winn and Shotton \citep{winn2006layout} proposed a LayoutCRF to model several occlusions for segmenting partially occluded objects. Gould \etal \citep{gould2009decomposing} decomposed a scene into semantic regions together with their spatial relationships. Sun \etal \citep{sun2010layered} utilized an MRF to model the layered image motion with occlusion ordering. Yang \etal~\citep{yang2010layered,yang2011layered} formulated a layered object detection and segmentation model, in which occlusion ordering for all detected objects was derived. This inferred order for all objects has been used to improve scene parsing~\citep{tighe2014scene} through a CRF. Zhang \etal \citep{zhang2015monocular} combined CNN and MRF to predict instance segmentation with depth ordering. While these methods evaluate occlusion ordering, their main goal is to improve the inmodal perception accuracy for object detection, image parsing, or instance segmentation using the spatial occlusion information. In contrast to these methods, our method \emph{not} only focuses on visible regions with structural inmodal perception, but also tries to solve for amodal perception. \ie to learn \emph{what is behind the occlusion}.

\par\medskip\noindent\textbf{Amodal image/instance perception.} Some initial steps have been taken toward amodal perception, exploring the invisible regions. Guo and Hoiem \citep{guo2012beyond} investigated background segmentation map completion by learning relationships between occluders and background. Subsequently, \citep{liu2016layered} introduced the Occlusion-CRF to handle occlusions and complete occluded surfaces. Kar \etal \citep{kar2015amodal} focused on amodal bounding box completion, where the goal is to predict the intact extent of the bounding box. The common attribute in these earlier amodal perception works is using piecemeal representations of a scene, rather than a full decomposition that infers the amodal shapes for all objects. 

The success of advanced deep networks trained on large-scale annotated datasets has recently led to the ability to get more comprehensive representations of a scene. Instance segmentation \citep{pinheiro2015learning,dai2016instance,pinheiro2016learning,li2017fully} deal with detecting, localizing and segmenting all objects of a scene into individual instances. This task combines the classical object detection \citep{girshick2014rich,he2015spatial,girshick2015fast,ren2015faster} and semantic segmentation \citep{long2015fully,chen2017deeplab,badrinarayanan2017segnet}. However, these notable methods typically segment the scene only into visible regions, and do \emph{not} have an explicit structural representation of a scene. We believe a key reason is the lack of large-scale datasets with corresponding annotations for amodal perception and occlusion ordering. The widely used datasets, such as Pascal VOC 2012~\citep{everingham2010pascal}, NYU Depth v2~\citep{silberman2012indoor}, COCO \citep{lin2014microsoft}, KITTI \citep{geiger2012we}, and CityScapes \citep{cordts2016cityscapes}, contain only annotations for the visible instances, purely aiming for 2D inmodal perception.

To mitigate the lack of annotated datasets, Li \etal \citep{li2016amodal} presented a self-supervised approach by pasting occluders into an image. Although reasonable amodal segmentation results are shown, a quantitative comparison is unavailable due to the lack of ground-truth annotations for invisible parts. In more recent works, the completed masks for occluded parts are provided in COCOA~\citep{zhu2017semantic} and KINS~\citep{qi2019amodal}, which are respectively a subset of COCO~\citep{lin2014microsoft} and KITTI~\citep{geiger2012we}. However, their annotations for invisible parts are manually labeled, which is highly subjective~\citep{ehsani2018segan,zhan2020self}. Furthermore, these datasets are mainly used for the task of inferring amodal semantic maps and are not suitable for the task of RGB appearance completion, since the ground truth RGB appearance for occluded parts are not available. In contrast, we jointly address these two amodal perception tasks using our constructed CSD dataset.

\begin{figure*}[tb!]
	\centering
	\includegraphics[width=\linewidth, height=0.295\textheight]{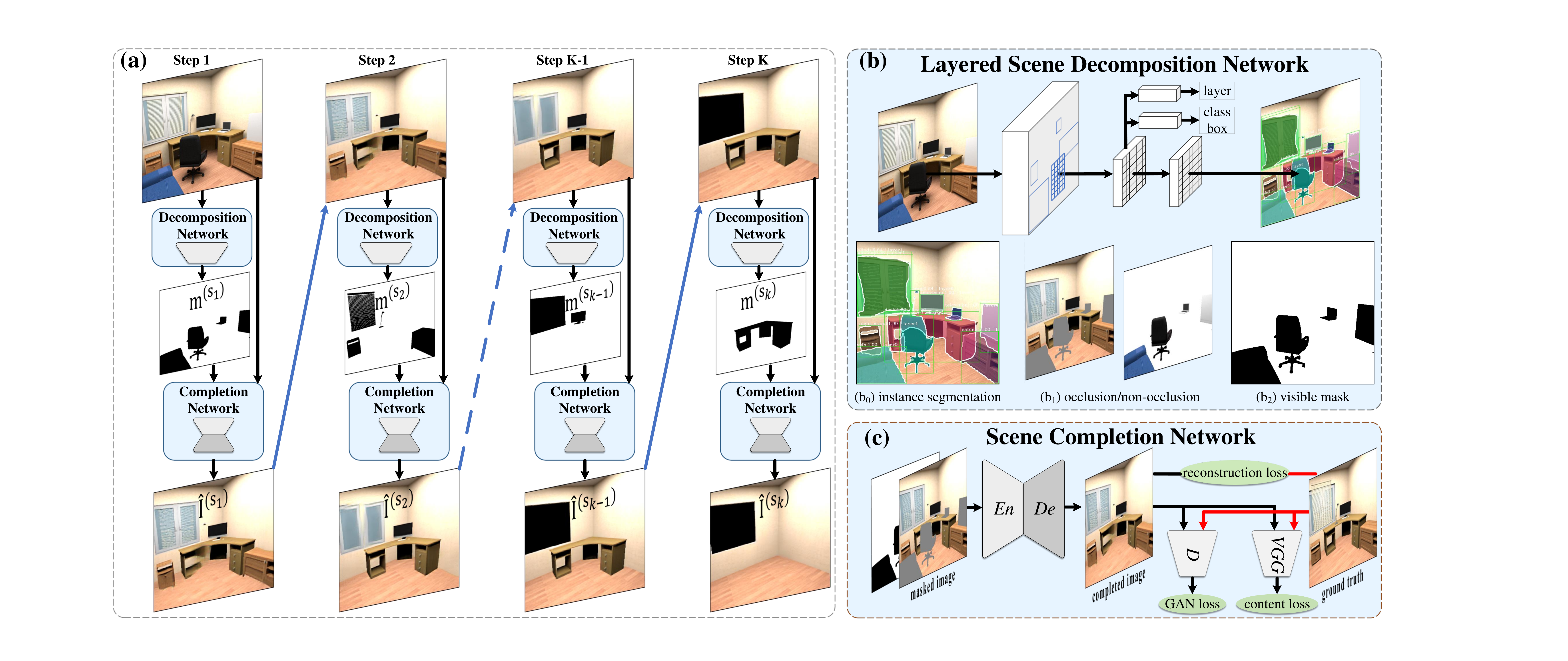}
	\caption{\textbf{An illustration of the CSDNet framework.} (a) Overall layer-by-layer completed scene decomposition pipeline. In each step, the layered decomposition network selects out the fully visible objects. The completion network will complete the resultant holes with appropriate imagery. The next step starts again with the completed image. (b) The layered scene decomposition network estimates instance masks and binary occlusion relationships. (c) The completion network generates realistic content for the original \emph{invisible} regions.}
	\label{fig:framework}
\end{figure*}

\par\medskip\noindent\textbf{Amodal perception for both mask and appearance.} The problem of generating the amodal RGB appearance for the occluded parts is highly related to semantic image completion. The latest methods \citep{pathak2016context,yang2017high,iizuka2017globally,yu2018generative,zheng2019pluralistic,nazeri2019edgeconnect} extend GANs \citep{goodfellow2014generative} and CGANs \citep{mirza2014conditional} to address this task, generating new imagery to fill in partially erased image regions. However, they mainly focus on object removal, needing users to interactively annotate the objects to be removed. 

SeGAN \citep{ehsani2018segan} involved an end-to-end network that sequentially infers amodal masks and generates complete RGB appearances for instances. The instance depth order is estimated by comparing the areas of the full and visible masks. PCNet~\citep{zhan2020self}  used a self-supervised learning approach to recover masks and content using only visible annotations. However, these works mainly present results in which the ground truth visible mask is used as input, and are sensitive to errors in this visible mask. As stated in~\citep{zhan2020self}, their focus is on amodal completion, rather than a scene understanding for amodal perception. While the recent work of Yan \etal \citep{yan2019visualizing} tried to visualize the invisible from a single input image, it only tackles the occluded ``vehicle'' category, for which there is much less variation in amodal shape and RGB appearance, and thus easier to model. 

There are two recent works that attempt to learn structural scene decomposition with amodal perception. MONet \citep{burgess2019monet} combined an attention network and a CVAE~\citep{kingma2013auto} for jointly modeling objects in a scene. While it is nominally able to do object appearance completion, this unsupervised method has only been shown to work on simple toy examples with minimal occlusions. Dhamo \etal \citep{Dhamo2019iccv} utilized Mask-RCNN \citep{he2017mask} to obtain visible masks, and conducted RGBA-D completion for each object. However, depth values are hard to accurately estimate from a single image, especially in real images without paired depth ground-truths. Besides, they still considered the decomposition and completion separately. In practice, even if we use domain transfer learning for depth estimation, the pixel-level depth value for all objects are unlikely to be consistent in a real scene. Therefore, our method uses an instance-level occlusion order, called the ``2.1D'' model \citep{yang2011layered}, to represent the structural information of a scene, which is easier to be inferred and manipulated.

\section{Method}\label{sec:method}

In this work, we aim to derive a higher-level structural decomposition of a scene. When given a single RGB image $\mathbf{I}$, our goal is to decompose all objects in it and infer their fully completed RGB appearances, together with their underlying occlusion relationships (As depicted in Fig.~\ref{fig:example}). Our system is designed to carry out inmodal perception for \emph{visible} structured instance segmentation, and also solve the amodal perception task of completing shapes and appearances for originally \emph{invisible} parts.

Instead of directly predicting the invisible content and decoupling the occlusion relationships of all objects at one pass, we use the fact that foreground objects are more easily identified, detected and segmented without occlusion. Our CSDNet decomposes the scene layer-by-layer. As shown in Fig.~\ref{fig:framework}(a), in each step $s_{k-1}$, given an image $\mathbf{I}^{(s_{k-1})}$, the layered segmentation network creates masks as well as occlusion labels for all detected objects. Those instances classified as fully visible are extracted out and the scene completion network generates appropriate appearances for the invisible regions. The completed image $\mathbf{\hat{I}}^{(s_{k-1})}$ will then be resubmitted for layered instance segmentation in the next step $s_{k}$. This differs significantly from previous works~\citep{ehsani2018segan,Dhamo2019iccv,burgess2019monet,zhan2020self,ling2020variational}, which do not adapt the segmentation process based on completion results.

Our \emph{key novel insight} is that scene completion generates completed shapes for originally occluded objects by leveraging the global scene context, so that they are subsequently easier to be detected and segmented without occlusion. Conversely, better segmented masks are the cornerstones to complete individual objects by precisely predicting which regions are occluded. Furthermore, this interleaved process enables extensive \emph{information sharing between these two networks}, to holistically solve for multiple objects, and produces a structured representation for a scene. This contrasts with existing one-pass methods~\citep{ehsani2018segan,Dhamo2019iccv,burgess2019monet,zhan2020self,ling2020variational}, where the segmentation and completion are processed separately and instances are handled independently. Together with the benefit of occlusion reasoning, our system is able to explicitly learn \emph{which parts of the objects and background are occluded that need to be completed}, instead of freely extending to arbitrary shapes.

\subsection{Layered Scene Decomposition}\label{sec:layered_scene}

As shown in Fig.~\ref{fig:framework}, our layered scene decomposition network comprehensively detect objects in a scene. For each candidate instance, it outputs a class label, a bounding-box offset and an instance mask. The system is an extension of Mask-RCNN \citep{he2017mask}, which consists of two main stages. In the first stage, the image is passed to a \emph{backbone network} (\eg ResNet-50-FPN \citep{lin2017feature}) and next to a \emph{region proposal network} (RPN \citep{ren2015faster}) to get object proposals. In the second stage, the network extracts features using \emph{RoIAlign} from each candidate box, for passing to object classification and mask prediction. We refer readers to \citep{he2017mask,chen2019hybrid} for details.

To determine if an object is fully visible \emph{or} partially occluded, a new parallel branch for this binary occlusion classification is added, as shown in Fig.~\ref{fig:framework}(b). This decomposition is done consecutively layer-by-layer, where at each step it is applied to a single RGB derived from the counterpart scene completion step. While only binary decisions are made in each step, after a full set of iterations, a comprehensive layered occlusion ordering is obtained. The following parts describe how this is done, looking first at the instance depth order representation, followed by how occlusion head is designed.

\par
\begin{figure}[tb!]
	\centering
	\includegraphics[width=\linewidth]{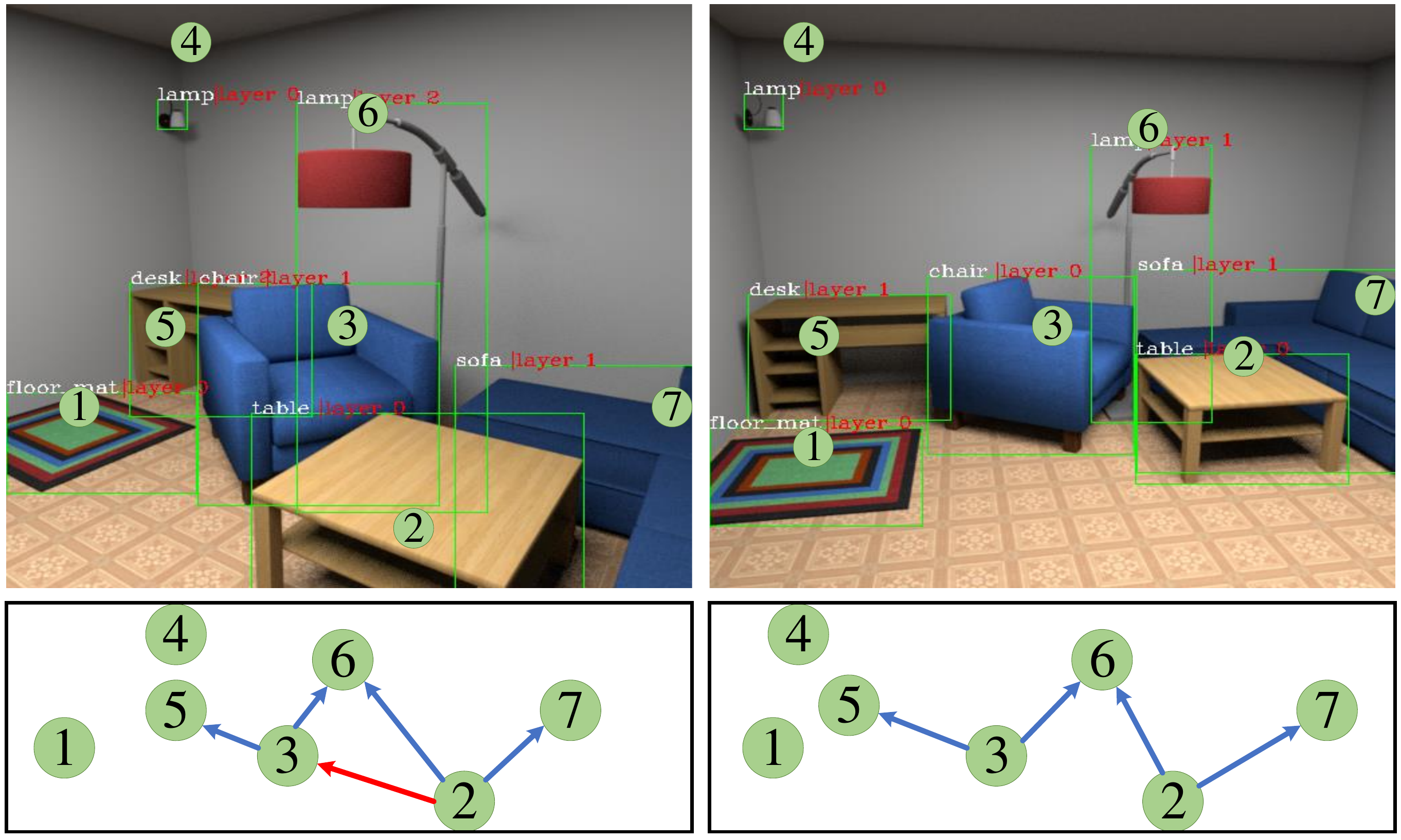}
	\begin{picture}(0,0)
	\put(-79,0){\footnotesize (a) View 1}
	\put(45,0){\footnotesize (b) View 2}
	\end{picture}
	\caption{\textbf{Instance depth order representation.} Top images show absolute layer order~\citep{qi2019amodal} in different views. Bottom directed graphs give the pairwise order between objects. }
	\label{fig:layer_representation}
\end{figure}
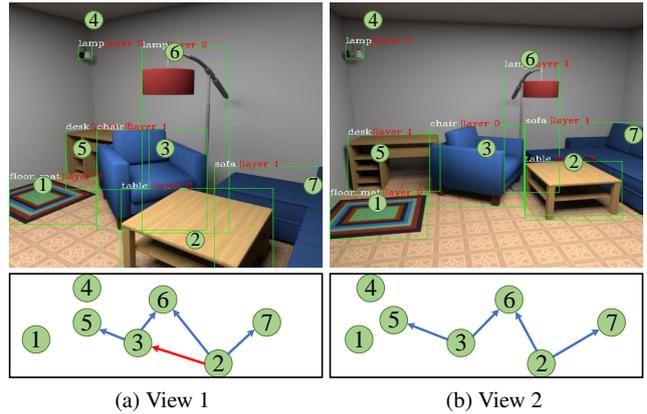

\medskip\noindent\textbf{Instance depth order representation.} Absolute layer order and pairwise occlusion order are two standard representations for occlusion reasoning in a scene \citep{sun2010layered,tighe2014scene}. As shown in Fig.~\ref{fig:layer_representation}, the definition for our \emph{absolute layer order} $\mathcal{L}$ follows~\citep{qi2019amodal}, where fully visible objects are labeled as 0, while other occluded objects have 1 order higher than the highest-order instance occluding them (see top images in Fig.~\ref{fig:layer_representation}). We interpret the \emph{pairwise occlusion order matrix} as a directed graph $G=(\Omega,W)$ (see bottom graphs in Fig.~\ref{fig:layer_representation}), where $\Omega$ is a discrete set of all instances with number $N$, and $W$ is a $N\times N$ matrix. $W_{i,j}$ is the occlusion relationship of instance $i$ to instance $j$. We use three numbers to encode the order --- $\{-1$: occluded, 0 : no relationship, 1: front$\}$. For example, the chair (instance \#3) is occluded by the table (instance \#2), so the pairwise order for the chair is $W_{3, 2} = -1$, while the pairwise order for the table is inversely labeled as $W_{2, 3} = 1$. 

\par\medskip\noindent\textbf{Occlusion head.} In practice we find it hard to directly predict these instance depth orders. The absolute layer order index $l\in\mathcal{L}$ cannot be predicted purely from local features in a bounding box, since it depends on the global layout of all objects in a scene. Furthermore, this index is very sensitive to viewpoints, \eg in Fig.~\ref{fig:layer_representation}, the desk (instance \#5) is occluded by only one object (instance \#3) in both views, but the absolute layer order indices of the desk are different: ``2'' \emph{vs} ``1''. In contrast, pairwise ordering $G$ captures the occlusion relationships between pairs of objects, but all pairs have to be analyzed, leading to scalability issues in the current instance segmentation network. As R-CNN-based system creates 2,000 candidate objects, this pairwise analysis requires building an unwieldy 2k$\times$2k features. %Even if we were to restrict these to the 100 highest scoring detection boxes, it will still be very memory intensive.

We circumvent these problems as our occlusion classifier only predicts a \emph{binary occlusion label}: $\{0, 1\}$ in each step, where $0$ is fully visible, and $1$ is occluded, following the setting of absolute layer order. During training, each ground-truth binary occlusion label is determined from the pairwise order of the actual objects present in the scene (see details in the Appendix). The occlusion head in our layered scene decomposition network is a \emph{fc} layer, which receives aligned features from each RoI and predicts the binary occlusion label. %Note that unlike the classification and bounding-box regression components that use shared features, our occlusion head maintains an individual branch to avoid corrupting the original object classification branch. 

\par\medskip\noindent\textbf{Decomposition Loss.} The multi-task loss function for layered scene decomposition is defined as follows:
\begin{equation}
{L}_\text{decomp} = \sum_{t=1}^{T}\alpha_t({L}_\text{cls}^t + {L}_\text{bbox}^t + {L}_\text{mask}^t+ {L}_\text{occ}^t) + \beta{L}_\text{seg}
\label{eq:decomposition_loss}
\end{equation}
where classification loss ${L}_\text{cls}^t$, bounding-box loss ${L}_\text{bbox}^t$, mask loss ${L}_\text{mask}^t$ and semantic segmentation loss ${L}_\text{seg}$ are identical to those defined in HTC \citep{chen2019hybrid}, and ${L}_\text{occ}^t$ is the occlusion loss at the cascade refined stage $t$ (three cascade refined blocks in HTC~\citep{chen2019hybrid}), using binary cross-entropy loss~\citep{long2015fully} for each RoI.

\subsection{Visiting the Invisible by Exploring Global Context}\label{sec:viv_global}

In our solution, we treat visiting the invisible as a \emph{semantic image completion}~\citep{pathak2016context} problem. As illustrated in Fig.~\ref{fig:framework}, in step $s_{k-1}$, after removing the front visible instances, the given image $\mathbf{I}^{(s_{k-1})}$ is degraded to become $\mathbf{I}_m^{(s_{k-1})}$. Our goal is to generate appropriate content to complete these previously \emph{invisible} regions (being occluded) for the next layer $\mathbf{I}^{(s_{k})}$. Unlike existing methods that complete each object independently~\citep{ehsani2018segan,Dhamo2019iccv,yan2019visualizing,burgess2019monet,zhan2020self,ling2020variational}, our model completes multiple objects in each step layer-by-layer, such that the information from earlier scene completions propagate to later ones. The global scene context is utilized in each step. 

To visit the invisible, it is critical to know which parts are invisible that need to be completed. The general image completion methods use manually interactive masks as input, which differs from our goal. Recent related works~\citep{ehsani2018segan,zhan2020self,ling2020variational} depend on the ground-truth visible masks as input to indicate which parts are occluded. In contrast, our system selects out fully visible objects and automatically learns which parts are occluded in each step. The holes left behind explicitly define the occluded regions for remaining objects, and thus the completed shapes for remaining objects must be deliberately \emph{restricted to these regions}, instead of being allowed to grow freely using only the predicted visible masks.

We use the PICNet~\citep{zheng2019pluralistic} framework to train our completion network. While the original PICNet was designed for diversity, here we only want to obtain the best result closest to the ground-truth. Therefore, we only use the encoder-decoder structure, and eschew the random sampling aspect.

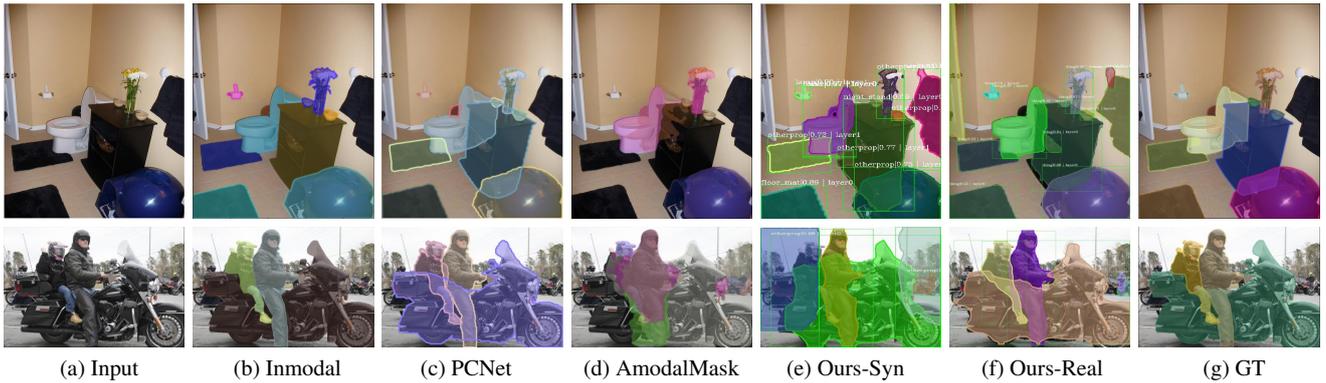
\begin{figure*}[htb!]
	\centering
	\includegraphics[width=\linewidth]{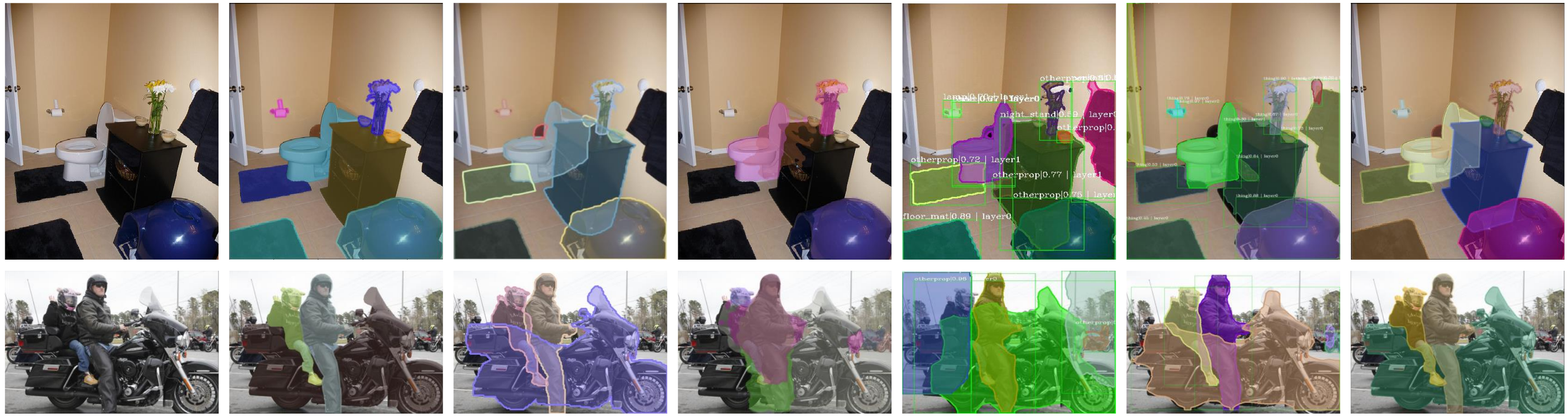}
	\begin{picture}(0,0)
	\put(-225,0){\footnotesize(a) Input}
	\put(-160,0){\footnotesize(b) Inmodal}
	\put(-90,0){\footnotesize(c) PCNet}
	\put(-29,0){\footnotesize(d) AmodalMask}
	\put(47,0){\footnotesize(e) Ours-Syn}
	\put(120,0){\footnotesize(f) Ours-Real}
	\put(203,0){\footnotesize(g) GT}
	\end{picture}
	\caption{\textbf{Amodal instance segmentation results on the COCOA validation set.} Our model trained on synthetic dataset (Ours-syn) achieves visually reasonable results in the similar real indoor scenes (example in top row), but it fails in some dissimilar real scenes (example in bottom row). After training on the real data with ``pseudo ground-truths'', the model (Ours-Real) performs much better. Note that, unlike the PCNet~\citep{zhan2020self} that need visible inmodal ground-truth masks as input, our system decomposes a scene using only a RGB image.}
	\label{fig:example_amodal_seg_real}
	\vspace{-0.1cm}
\end{figure*}

%\par\medskip\noindent\textbf{Patch Attention.} As most objects contain self-similar repeated textures, transferring such textures from visible to invisible regions is important for high-quality completion. While the PICNet uses a short-long term attention layer to copy the visible information to missing holes, the attention score is based on single points. Inspired by PatchMatch \cite{barnes2009patchmatch}, we use the mean attention score of a patch (\eg $3\times3$) rather than the usual single point attention in \cite{zheng2019pluralistic}. Formally, given the previous decoder features $\mathbf{f}_d$, we first calculate the point attention score of:
%\begin{equation}
%\beta_{j,i} = \frac{\exp(s_{ij})}{\sum_{i=1}^{N}\exp(s_{ij})},\mbox{where } s_{ij}=Q(f_{di})^TQ(f_{dj}),
%\end{equation}
%where $\beta_{j,i}$ represents the similarity of the $i^{th}$ to the $j^{th}$ location in the decoder features. After obtaining the self-attention score $\beta_{j,i}$ for each pixel, we further ensure the consistency of attention maps by fusing the similarity score in a square patch: $\hat\beta_{j,i} = \sum_{{j}'\in U_j,{i}'\in U_i}\beta_{{j}',{i}'}$, where $U_j$ and $U_i$ are the neighborhood patch sets at $j^{th}$ and $i^{th}$ locations separately. This attention map facilitates transfer of features from visible to originally occluded regions. 

\par\medskip\noindent\textbf{Completion Loss.} The overall scene completion loss function is given by
\begin{equation}
{L}_\text{comp} = \alpha_\text{rec}{L}_\text{rec} + \alpha_\text{ad}{L}_\text{ad} + \alpha_\text{per}{L}_\text{per}
\label{eq:completion_loss}
\end{equation}
where reconstruction loss ${L}_\text{rec}$ and adversarial loss ${L}_\text{ad}$ are identical to those in PICNet \citep{zheng2019pluralistic}. The perceptual loss ${L}_\text{per}=|\mathbf{F}^{(l)}(\mathbf{\hat{I}}^{(s_{k})}) - \mathbf{F}^{(l)}(\mathbf{I}^{(s_{k})})|$ \citep{johnson2016perceptual}, based on a pretrained VGG-19 \citep{simonyan2014very}, is the $l_1$ distance of features $\mathbf{F}$ in $l$-th layer between the generated image $\mathbf{\hat{I}}^{(s_{k})}$ and ground-truth $\mathbf{I}^{(s_{k})}$.

\subsection{Inferring Instance Pairwise Occlusion Order}\label{sec:infer_order}

As discussed in Section~\ref{sec:layered_scene}, absolute layer order $\mathcal{L}$ is sensitive to errors. If one object is incorrectly selected as a front object in an earlier step, objects behind it will have their absolute layer order incorrectly shifted. Hence in keeping with prior works~\citep{ehsani2018segan,zhan2020self}, we use the pairwise occlusion order $G=(\Omega,W)$ to represent our final instance occlusion relationships for evaluation. 

Given a single image $\mathbf{I}$, our model decomposes it into instances with completed RGB appearances $A_\Omega^{S_K}$. Here, $A$ denotes the amodal perception instance (inclusive of both mask and appearance), $\Omega$ specifies instances in the scene, and $S_K$ indicates which layers are the instances in (selected out in step $s_k$). When two segmented amodal masks $A_{\omega_i}^{s_i}$ and $A_{\omega_j}^{s_j}$ overlap, we infer their occlusion relationship based on the order of the object-removing process, formally:
\begin{equation}
W_{\omega_i,\omega_j} = \left\{\begin{array}{ll}
0 & \textrm{if $O(A_{\omega_i}^{s_i}, A_{\omega_j}^{s_j})=0$}\\
1 & \textrm{if $O(A_{\omega_i}^{s_i}, A_{\omega_j}^{s_j})>0$ and $s_i<s_j$}\\
-1 & \textrm{if $O(A_{\omega_i}^{s_i}, A_{\omega_j}^{s_j})>0$ and $s_i\geq s_j$}\\
\end{array} \right.
\end{equation}
where $O(A_{\omega_i}^{s_i}, A_{\omega_j}^{s_j})$ is the area of overlap between instances $\omega_i$ and $\omega_j$. If they do not overlap, they share no pairwise depth-order relationship in a scene. If there is an overlap and the instance $\omega_i$ is first selected out with a smaller layer order, the inferred pairwise order is $W_{\omega_i,\omega_j}$ = 1; otherwise it is labeled as $W_{\omega_i,\omega_j}$ = -1. Hence the instance occlusion order only depends on the order (selected out step) of removal between the two instances, and do not suffer from shift errors. 

\begin{figure*}
	\begin{minipage}[c]{0.48\textwidth}
			\centering
			\includegraphics[width=\linewidth,height=0.20\textheight]{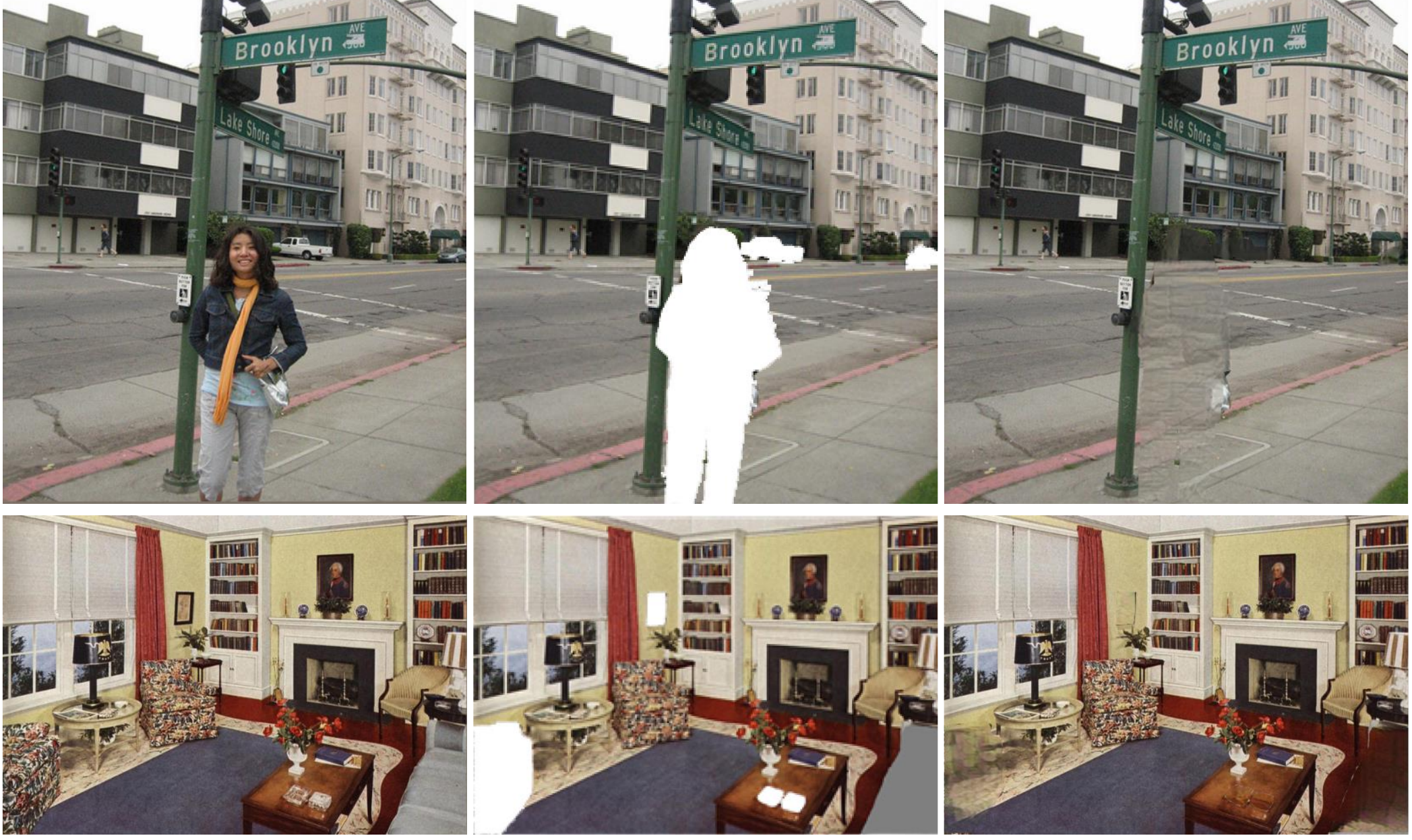}
			\begin{picture}(0,0)
			\put(-96,0){\footnotesize (a) Input}
			\put(-32,0){\footnotesize (b) Masked Image}
			\put(44,0){\footnotesize (c) Completed Image}
			\end{picture}
			\caption{\textbf{Pseudo RGB ground-truth.} (a) Input. (b) Masked image by selecting out the fully visible objects. (c) Pseudo ground-truth generated from our model trained on synthetic data.}
			\label{fig:syn_completed}
	\end{minipage}
	\hspace{0.02\linewidth}
    \begin{minipage}[c]{0.48\textwidth}
    	\centering
    	\includegraphics[width=\linewidth]{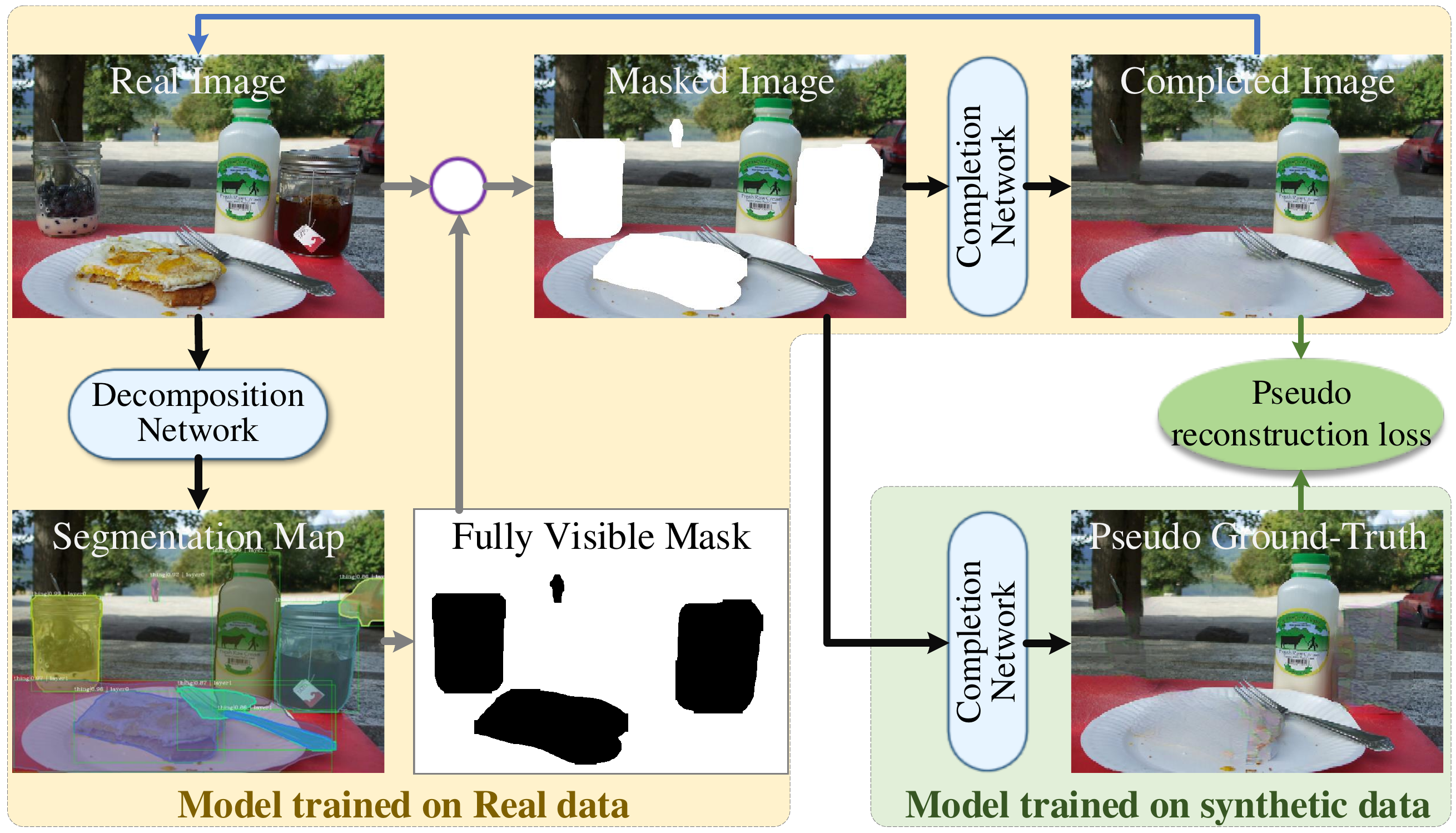}
    	\caption{\textbf{Training pipeline for real images.} We introduce a semi-supervised learning method for real data by providing \emph{pseudo} RGB ground-truth for originally invisible regions.}
    	\label{fig:framework_real}
    \end{minipage}
\end{figure*}

\subsection{Training on Real Data with Pseudo Ground-truth}

Real-world data appropriate for a completed scene decomposition task is difficult to acquire, because ground truth shapes and RGB appearance for occluded parts are hard to collect without very extensive manual interventions, \eg deliberate physical placement and removal of objects. Although our proposed model trained on the high-quality rendered data achieves visually reasonable results in some real scenes that share similarities to the rendered dataset (\eg indoor scene in top row of Fig.~\ref{fig:example_amodal_seg_real}), it does not generalize well to dissimilar real scenes (\eg outdoor scene in bottom row of Fig.~\ref{fig:example_amodal_seg_real}). These are caused by: 1) differences in labeled categories between synthetic and real datasets, especially between indoor and outdoor scenes; and 2) inconsistencies between synthetically trained image completion of masked regions and fully visible real pixels.

One alternative is to simply use an image completion network trained only on real images. From our experience, this performs poorly in a scene decomposition task. The reason is that while real-trained image completion methods are able to create perceptually-pleasing regions and textures for a single completion, they do not appear to have the ability to adhere to consistent object geometry and boundaries when de-occluding, which is crucial for object shape completion. As a result, errors accumulate even more dramatically as the decomposition progresses.

Our \emph{key motivating insight} is this: instead of training the model entirely without ground-truth in the completion task, we train it in a semi-supervised learning manner, exploiting the scene structure and object shape knowledge that has been gained in our synthetically-trained CSDNet. As shown in Fig.~\ref{fig:syn_completed}, this synthetic completion model is able to generate visually adequate appearance, but more importantly it is better able to retain appropriate geometry and shapes. We can use this to guide the main image completion process in real-word data, while allowing a GAN-based loss to increase the realism of the output.

Specifically, for a real image $\mathbf{I}$, we first train the layered decomposition network using the manual annotated amodal labels. In a step, after segmenting and selecting out the foreground objects, we obtain $\mathbf{\hat{I}}_{syn}^{(s_k)}=G(\mathbf{I}_m^{(s_k)};\theta_{\text{syn}})$ to serve as ``pseudo ground-truth'' (green box in Fig.~\ref{fig:framework_real}) through the completion model trained on synthetic data. We then train the completion network $G(\mathbf{I}_m^{(s_k)};\theta_{\text{real}})$ using the loss function of quation (\ref{eq:completion_loss}) by comparing the output $\mathbf{\hat{I}}_{real}^{(s_k)}$ to ``pseudo ground-truth'' $\mathbf{\hat{I}}_{syn}^{(s_k)}$. Like \citep{sengupta2020background}, we also reduce the weights of reconstruction loss ${L}_\text{rec}$ and perceptual loss ${L}_\text{per}$ to encourage the output to be biased towards the real image distribution via the discriminator loss ${L}_\text{ad}$. It is worth noticing that the completed image is \emph{passed back} to the layered decomposition network in the next layer, where the decomposition loss ${L}_\text{decomp}$ in equation (\ref{eq:decomposition_loss}) will be backpropagated to the completion network. This connection allows the completion network to \emph{learn to complete real world objects that might not be learned through the synthetic data}.

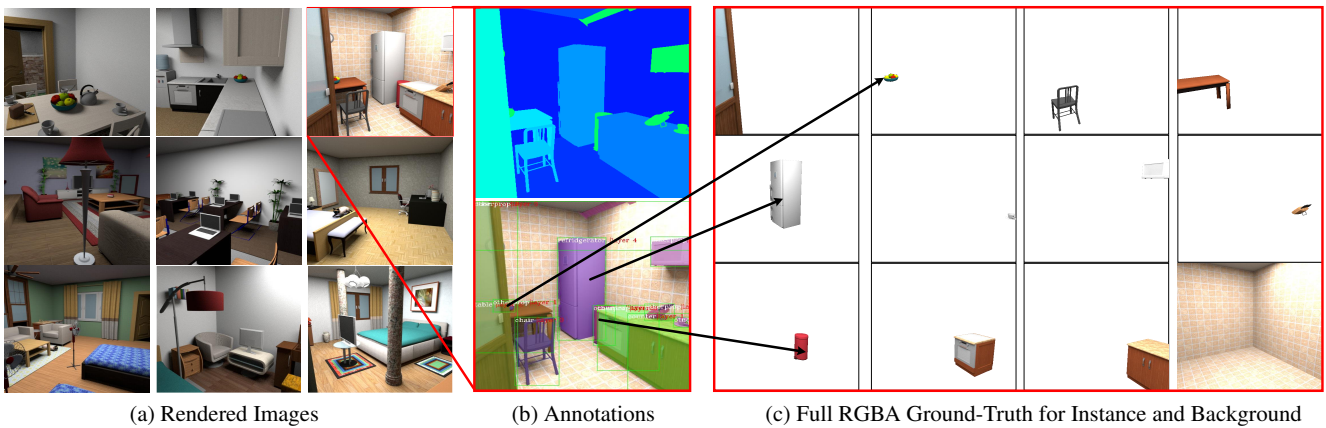
\begin{figure*}[tb!]
	\centering
	\includegraphics[width=\linewidth,height=0.22\textheight]{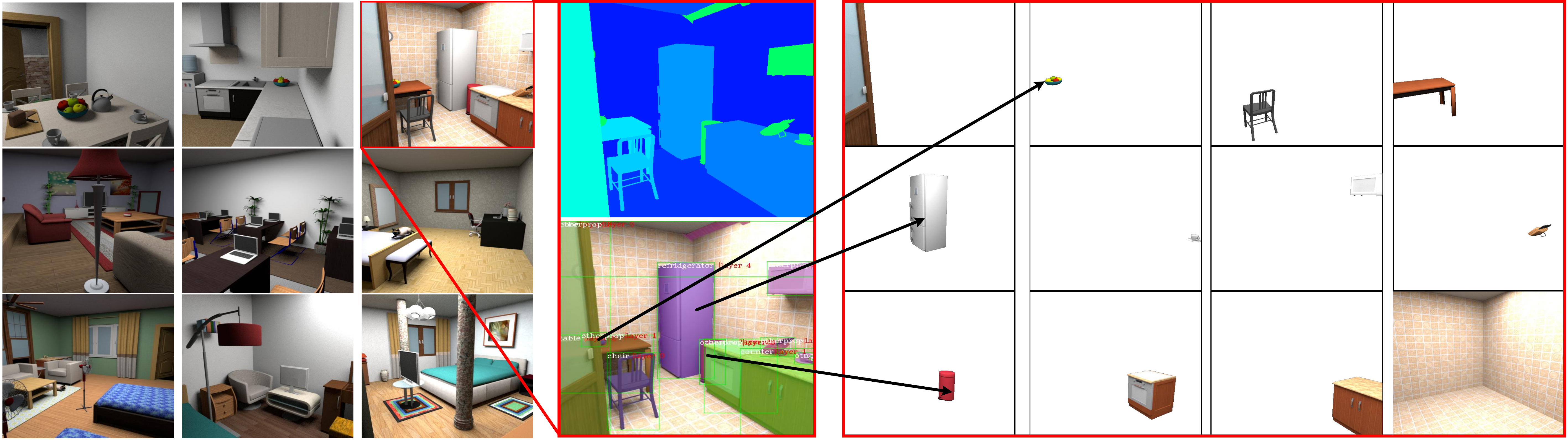}
	\begin{picture}(0,0)
	\put(-200,0){\footnotesize(a) Rendered Images}
	\put(-57,0){\footnotesize(b) Annotations}
	\put(38,0){\footnotesize(c) Full RGBA Ground-Truth for Instance and Background}
	\end{picture}
	\caption{\textbf{Our rendered dataset.} (a) High quality rendered RGB images. (b) Semantic map and instance annotations with bbox, category, ordering and segmentation map. (c) Intact RGBA ground-truth for instances and background.}
	\label{fig:datasets}
\end{figure*}

\section{Synthetic Data Creation}\label{sec:data}

Large datasets with complete ground-truth appearances for all objects are very limited. Burgess \etal \citep{burgess2019monet} created the \emph{Objects Room dataset}, but only with toy objects. Ehsani \etal \citep{ehsani2018segan} and Dhamo \etal \citep{Dhamo2019iccv} rendered synthetic datasets. However, the former only includes 11 rooms, with most viewpoints being atypical of real indoor images. The latter's OpenGL-rendered dataset appears to have more typical viewpoints with rich annotations, but the OpenGL-rendered images have low realism. Recently, Zhan \etal \citep{zhan2020self} explored the \emph{amodal completion} task through self-supervised learning without the need of amodal ground-truth. However, a fair quantitative comparison is not possible as no appearance ground-truth is available for invisible parts. 

To mitigate this issue, we rendered a realistic dataset with Maya \citep{Maya}. We can train the supervised model and test the unsupervised model on this synthetic data with masks and RGB appearance ground-truths for all occluded parts. 

\par\medskip\noindent\textbf{Data Rendering.} Our rendered data is based on a total of 10.2k views inside over 2k rooms (CAD models from SUNCG \citep{song2017semantic}) with various room types and lighting environments (see Fig.~\ref{fig:datasets}(a)). To select the typical viewpoints, we first sampled many random positions, orientations and heights for the camera. Only when a view contains at least 5 objects will we render the image and the corresponding ground-truth of each instance. To avoid an excessive rendering workload, we separately rendered each isolated object, as well as the empty room, as shown in Fig.~\ref{fig:datasets}(c). This allows us to then freely create the ground-truths of each layer by compositing these individual objects and background using the instance occlusion order. The rendering details and statistics of the dataset can be found in the Appendix. 

\par\medskip\noindent\textbf{Data Annotation.} Each rendered scene is accompanied by a global semantic map and dense annotations for all objects. As shown in Fig.~\ref{fig:datasets}(b) and Fig.~\ref{fig:datasets}(c), the intact RGB appearances are given, as well as categories (the 40 classes in NYUDv2 \citep{Silberman:ECCV12}), bounding boxes and masks for complete objects, as well as for only the visible regions. Furthermore, the absolute layer order and pairwise occlusion order shown in Fig.~\ref{fig:layer_representation} are also defined in our rendered dataset.

\begin{figure*}[tb!]
	\centering
	\includegraphics[width=\linewidth,height=0.38\textheight]{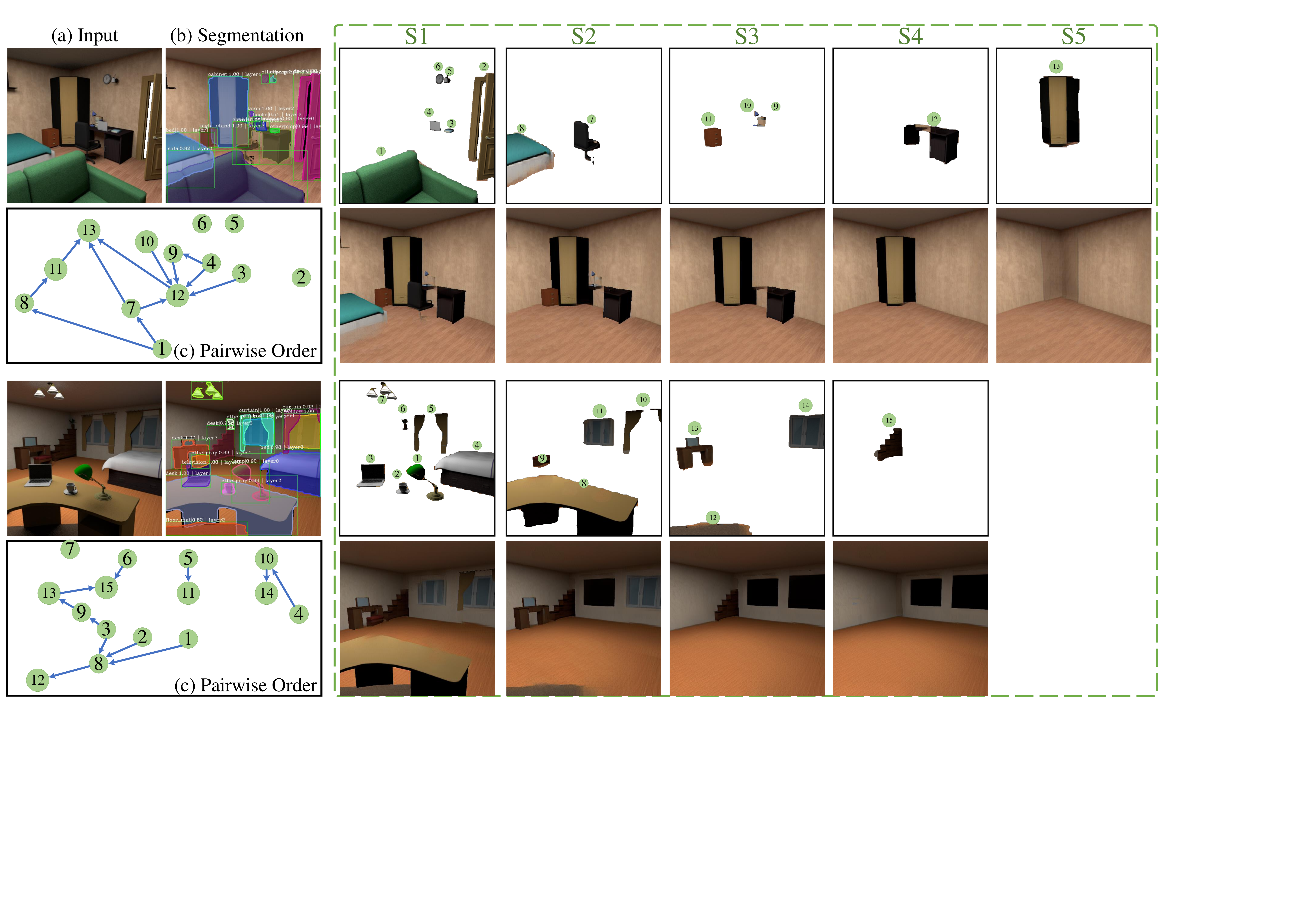}
	\caption{\textbf{Layer-by-Layer Completed Scene Decomposition} results on rendered CSD testing set. (a) Input RGB images. (b) Final amodal instance segmentation. (c) Inferred directed graph for pairwise order. (d) Columns labeled S1-5 show the decomposed instances (top) and completed scene (bottom) based on the predicted non-occluded masks. Note that the originally occluded invisible parts are filled in with realistic appearance. }
	\label{fig:results_de_co}
\end{figure*}

\section{Experiments}\label{sec:experiment}

\subsection{Setup}

\par\medskip\noindent\textbf{Datasets.} We evaluated our system on three datasets: \textbf{COCOA}~\citep{zhu2017semantic}, \textbf{KINS}~\citep{qi2019amodal} and the rendered \textbf{CSD}. \textbf{COCOA} is annotated from COCO2014~\citep{lin2014microsoft}, a large scale natural image datasets, in which 5,000 images are selected to manually label with pairwise occlusion orders and amodal masks. \textbf{KINS} is derived from the outdoor traffic dataset KITTI~\citep{geiger2012we}, in which 14,991 images were labeled with absolute layer orders and amodal masks. \textbf{CSD} is our rendered synthetic dataset, which contains 8,298 images, 95,030 instances for training and 1,012 images, 11,648 instances for testing. We conducted thorough experiments and ablation studies to assess the quality of the completed results for invisible appearance estimation (since the in-the-wild datasets lack ground-truth for the occluded parts).

\par\medskip\noindent\textbf{Metrics.} For amodal instance segmentation, we report the standard COCO metrics \citep{lin2014microsoft}, including AP (average over IoU thresholds), $\text{AP}_{50}$, $\text{AP}_{75}$, and $\text{AP}_{S}$, $\text{AP}_{M}$ and $\text{AP}_{L}$ (AP at different scales). Unless otherwise stated, the AP is for mask IoU. For appearance completion, we used RMSE, SSIM and PSNR to evaluate the quality of generated images. All images were normalized to the range $\left[0, 1\right]$.

Since the occlusion order is related to the quality of instance segmentation, we defined a novel metric for evaluating the occlusion order that uses the previous benchmark criterion for instance segmentation. Specifically, given a pairwise occlusion order $G=(\Omega, W)$ predicted by the model, we only evaluate the order for these valid instances that have IoU with ground-truth masks over a given threshold. For instance, if we set the threshold as $0.5$, the predicted instance $\omega$ will be evaluated when we can identify a matched ground-truth mask with $\text{IoU}\geq 0.5$. Hence we can measure the \textbf{occlusion average precision} (OAP) as assessed with different thresholds. 

\begin{table*}[tb!]
	\centering
	\caption{\textbf{Amodal Instance Segmentation on CSD testing sets}. Mask-RCNN~\citep{he2017mask} and HTC~\citep{chen2019hybrid} are the state-of-the-arts of the COCO segmentation challenges. MLC~\citep{qi2019amodal} is the latest amodal instance segmentation work for outdoor scene. PCNet~\citep{zhan2020self} is the self-supervised amodal completion work. The \textbf{CSDNet}-gt holds same training environment as \textbf{CSDNet}, but is tested with completed ground-truths images $\mathbf{I}^{s_*}$ in each step. Best results used ground-truth annotations are marked with $^*$, while best results only used RGB images are in bold.}
	\label{table:syn_seg}
	\renewcommand{\arraystretch}{1.1}
	\begin{tabular}{@{}l|l|c |c c c |c c c@{}}
		\hlineB{3}
		& SegNet & box AP & mask AP & AP$_{50}$ & AP$_{75}$ & AP$_{S}$ & AP$_M$& AP$_L$ \\
		\hlineB{3}
		Mask-RCNN \citep{he2017mask}& Mask-RCNN & 51.3 & 46.8 & 67.2 & 50.6 & 14.5 & 43.0 & 49.9 \\
		MLC \citep{qi2019amodal}& Mask-RCNN & 52.3 & 47.2 & 67.5 & 50.9 & 14.7 & 43.8 & 50.2 \\
		PCNet \citep{zhan2020self} & Mask-RCNN & - & 43.6 & 59.1 & 43.4 & 11.5 & 40.4 & 46.0 \\
		HTC \citep{chen2019hybrid} & HTC &52.9 & 47.3 & 65.9 & 51.6 & 12.2 & 41.3 & 51.0 \\
		MLC \citep{qi2019amodal}& HTC & 53.6 & 47.9 & 66.1 & 52.3 & 13.1 & 41.9 & 51.7 \\
		PCNet \citep{zhan2020self} & HTC & - & 45.7 & 60.6 & 49.2 & 10.2 & 39.3 & 48.4 \\
		\hline
		\textbf{CSDNet}& Mask-RCNN & 52.6  & 48.7  &  66.2 &  53.1 &  15.7 & 42.8 & 52.2 \\
		\textbf{CSDNet} & HTC &\bf{56.3} & \bf{50.3} & \bf{67.7} & \bf{53.4} & \bf{17.4} & \bf{44.2} & \bf{53.1} \\
		\hline
		\textbf{CSDNet}-gt& Mask-RCNN &  54.9 &  53.1 &  66.5 &  56.9 &  21.4$^*$ &  49.9 &  57.0 \\
		\textbf{CSDNet}-gt & HTC & 60.3$^*$ & 56.0$^*$ & 67.9$^*$ & 59.3$^*$ & 19.6 & 53.4$^*$ & 59.5$^*$ \\
		\hlineB{2.5}
	\end{tabular}
\end{table*}

\begin{table*}[tb!]
	\centering
	\caption{\textbf{Instance depth ordering on CSD testing sets}. We report the pairwise depth ordering on occluded instance pairs $\text{OAP}_{occ}$. I$^{s*}=$ ground-truth completed image in each step ${s*}$, $\text{V}_{gt}=$ visible ground-truth mask, $\hat{\text{V}}_{pred}=$ visible predicated mask, and $\hat{\text{F}}_{pre}=$ full (amodal) predicated mask. $\text{layer order}^1$ only predicts the occlusion / non-occlusion labels in the original image (the first step in our model).}
	\label{table:syn_depth}
	\renewcommand{\arraystretch}{1.1}
	\begin{tabular}{@{}l| l|l | l |c c c c|c c c@{}}
		\hlineB{3}
		& \multicolumn{2}{c|}{Inputs} & \multicolumn{1}{c|}{\multirow{2}{*}{\makecell[c]{Ordering \\ Algorithm}}} & \multirow{2}{*}{OAP} & \multirow{2}{*}{OAP$_{50}$} & \multirow{2}{*}{OAP$_{75}$} & \multirow{2}{*}{OAP$_{85}$} & \multirow{2}{*}{OAP$_{S}$} & \multirow{2}{*}{OAP$_M$}& \multirow{2}{*}{OAP$_L$} \\
		\cline{2-3}
		& \makecell[c]{Amodal} & \makecell[c]{Ordering} & & & & & & & & \\
		\hlineB{3}
		SeGAN \citep{ehsani2018segan} & I + $\text{V}_{gt}$ & $\text{V}_{gt}$ + $\hat{\text{F}}_{pre}$ & IoU Area & 68.4 & - & - & - & - & - & -\\
		SeGAN \citep{ehsani2018segan} & I + $\hat{\text{V}}_{pred}$ & $\text{V}_{gt}$ + $\hat{\text{F}}_{pre}$ & IoU Area & 66.1 & 50.2 & 65.6 & 70.4 & 10.6 & 65.0 & 63.8 \\
		HTC + MLC \citep{qi2019amodal} & I & $\text{V}_{gt}$ + $\hat{\text{F}}_{pre}$ & IoU Area &  76.5 &  70.3 &  77.1 &  79.8 &  11.6 &  69.8 &  78.2 \\
		HTC + MLC \citep{qi2019amodal} & I & $\hat{\text{F}}_{pre}$ + layer &layer order$^1$ & 51.9 & 44.3 & 50.8 & 54.6 & 11.7 & 60.8 & 46.2 \\
		HTC + PCNet \citep{zhan2020self} & I + $\hat{\text{V}}_{pred}$ & $\text{V}_{gt}$ + $\hat{\text{F}}_{pre}$ & IoU Area &  70.8 & 56.9 & 71.3 & 76.0 & 11.3 & 67.1 & 68.6 \\
		\hline
		\textbf{CSDNet} & I & $\hat{\text{F}}_{pre}$ & Area & 44.7 & 45.3 & 45.7 & 45.1 & 17.4 & 34.5 & 41.5 \\
		\textbf{CSDNet} & I & $\hat{\text{F}}_{pre}$ & Y-axis & 62.0 & 60.1 & 61.2 & 62.7 & \bf{63.4} & 58.6 & 66.1 \\
		\textbf{CSDNet} & I & $\text{V}_{gt}$ + $\hat{\text{F}}_{pre}$ & IoU Area & 80.7 & \bf{77.2} & \bf{81.0} & 82.9 & 61.1 & 73.7 & 80.5\\
		\textbf{CSDNet} & I & $\hat{\text{F}}_{pre}$ + layer & layer order & \bf{81.7} & 76.6 & 80.9 & \bf{84.6} & 15.7 & \bf{75.9} & \bf{82.6}\\
		\hline
		\textbf{CSDNet}-gt & I$^{s*}$ & $\hat{\text{F}}_{pre}$ + layer & layer order & 88.9$^*$ & 85.2$^*$ & 88.5$^*$ & 90.1$^*$ & 49.6 & 84.3$^*$ & 89.9$^*$ \\
		\hlineB{2.5}
	\end{tabular}
\end{table*}

\subsection{Results on Synthetic CSD Dataset}\label{sec:csd_results}

We first present results that we obtained from our framework when experimenting on our synthetic CSD dataset.

\subsubsection{Main Results}

\par\medskip\noindent\textbf{Completed scene decomposition.} We show the qualitative results of CSDNet in Fig.~\ref{fig:results_de_co}. Given a single RGB image, the system has learned to decompose it into semantically complete instances (\eg counter, table, window) and the background (wall, floor and ceiling), while completing RGB appearance for \emph{invisible} regions. Columns labeled S1-5 show the completed results layer-by-layer. In each layer, fully visible instances are segmented out, and after scene completion some previously occluded regions become fully visible in the next layer, \eg the table in the second example. The final amodal instance segmentation results shown in Fig.~\ref{fig:results_de_co}(b) consist of the fully visible amodal masks in each layer. Note that unlike MONet~\citep{burgess2019monet}, our model does not need predefined slots. The process will stop when it is unable to detect any more objects.

\par\medskip\noindent\textbf{Amodal instance segmentation.} We compare CSDNet to the state-of-the-art methods in amodal instance segmentation in  Table~\ref{table:syn_seg}. As the existing works Mask-RCNN~\citep{he2017mask} and HTC~\citep{chen2019hybrid} are aimed at inmodal perception for visible parts, we retrained their models for amodal perception task, by providing amodal ground-truths. We also retrained MLC~\citep{qi2019amodal} on our rendered dataset, which are the latest work for amodal perception. For PCNet~\citep{zhan2020self}, we used the predicted visible mask as input, rather than the original visible ground-truth annotations. While the HTC~\citep{chen2019hybrid} improves on Mask-RCNN's~\citep{he2017mask}  bounding box AP by about 1.6 points by refining the bounding box offsets in three cascade steps, the improvement for amodal mask segmentation is quite minor at 0.5 points. We believe this is an inherent limitation of methods that attempt amodal segmentation of occluded objects directly without first reasoning about occluding objects and masking their image features, as such the front objects' features will distract the network. In contrast, our CSDNet is able to improve the amodal mask segmentation accuracy by a relative $6.3\%$ with the same \emph{backbone segmentation network} (HTC), by jointing segmentation and completion with layer-by-layer decomposition.

To further demonstrate that better completed images improve amodal segmentation, we consider a scenario with a completion oracle, by using ground-truth appearances to repair the occluded parts in each step. This is denoted as the \textbf{CSDNet}-gt, for which amodal instance segmentation accuracy increases from $47.3\%$ to $56.0\%$ (relative $18.4\%$ improvement). We also note that, while the \textbf{CSDNet}-gt using Mask-RCNN achieves lower bounding box score than our HTC \textbf{CSDNet} (``54.9'' \emph{vs} ``56.3''), the mask accuracy is much higher (``53.1'' \emph{vs} ``50.3''). This suggests that amodal segmentation benefits from better completed images. 

\par\medskip\noindent\textbf{Instance depth ordering.} Following~\citep{zhu2017semantic}, we report the pairwise instance depth ordering for correctly detected instances in Table~\ref{table:syn_depth}. The original SeGAN and PCNet used ground-truth visible masks $\text{V}_{gt}$ as input. For a fair comparison, we first retrained them on our synthetic data using the same segmentation network (HTC~\citep{chen2019hybrid}) for all models. After predicting amodal masks, we assessed various instance depth ordering algorithms: two baselines proposed in AmodalMask~\citep{zhu2017semantic} of ordering by \emph{area} \footnote{We used the heuristic in PCNet~\citep{zhan2020self} --- larger masks are ordered in front for KINS, and behind for COCOA and CSD.} and by \emph{y-axis} (amodal masks closest to image bottom in front), ordering by \emph{incremental area} defined as the \emph{IoU area} between visible and amodal masks\footnote{See details in \citep{zhan2020self}, where the visible ground-truth masks $\text{V}_{gt}$ are used for ordering.}, and our ordering by absolute \emph{layer order} (Section~\ref{sec:infer_order}).

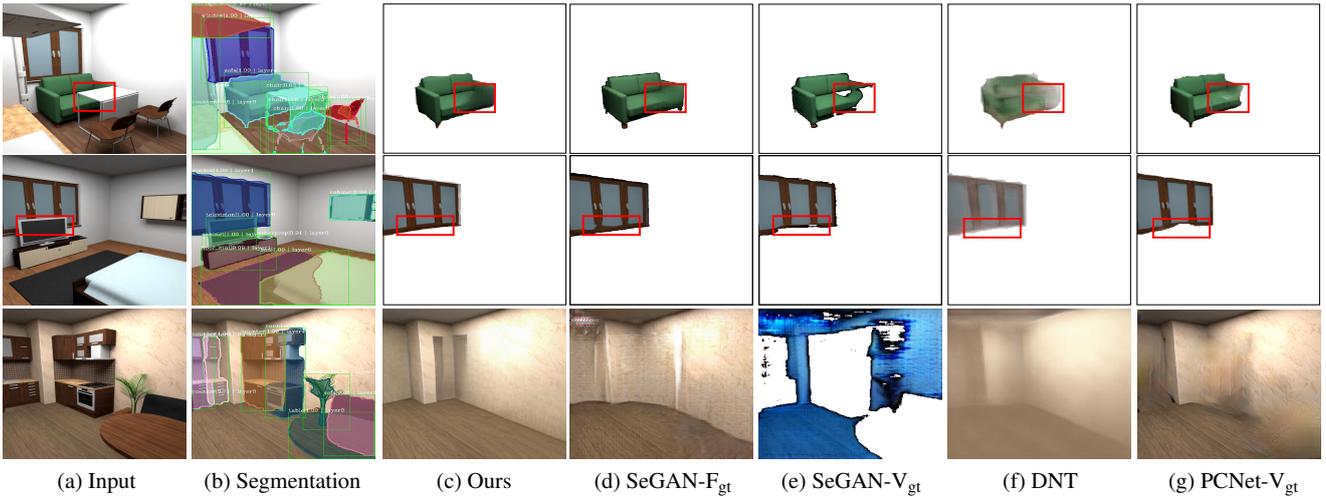
\begin{figure*}[tb!]
	\centering
	\includegraphics[width=\linewidth,height=0.26\textheight]{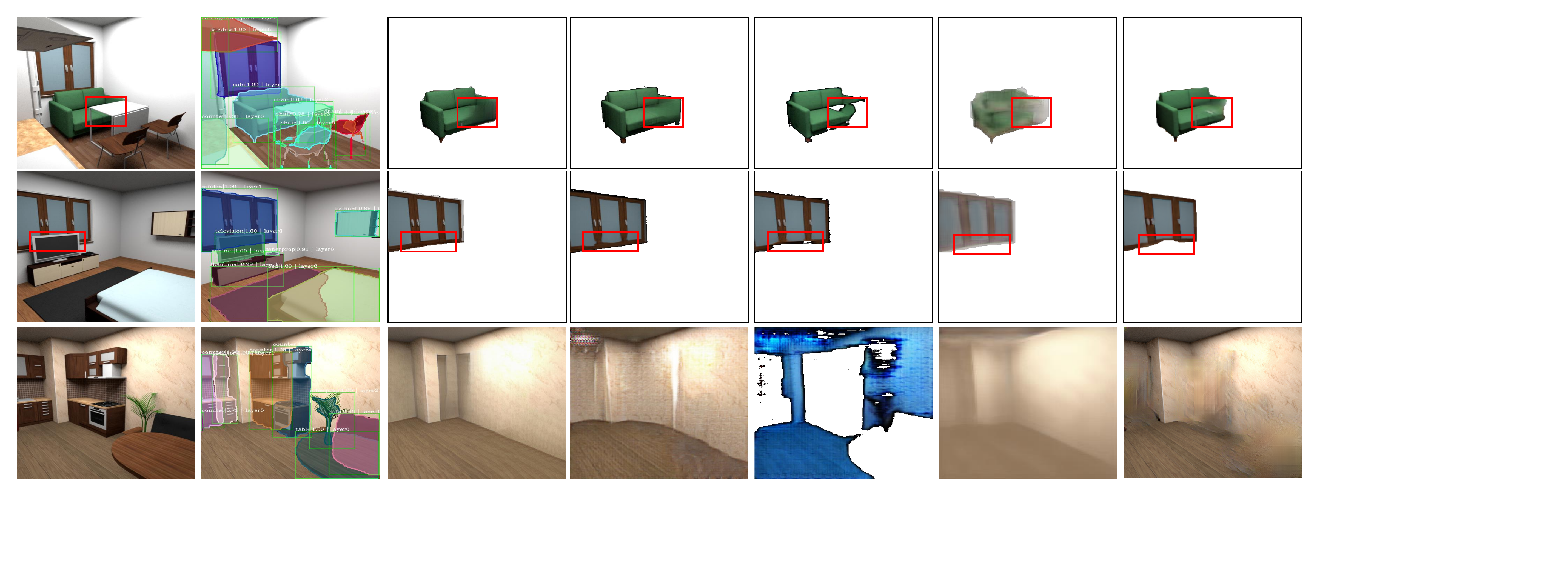}
	\begin{picture}(0,0)
	\put(-226,0){\footnotesize(a) Input}
	\put(-171,0){\footnotesize(b) Segmentation}
	\put(-84,0){\footnotesize(c) Ours}
	\put(-25,0){\footnotesize(d) SeGAN-F$_\text{gt}$}
	\put(45,0){\footnotesize(e) SeGAN-V$_\text{gt}$}
	\put(128,0){\footnotesize(f) DNT}
	\put(190,0){\footnotesize(g) PCNet-V$_\text{gt}$}
	\end{picture}
	\caption{Results for \textbf{Visiting the Invisible}. We show the input image, our amodal instance segmentation results, and the objects and background we try to visit. The red rectangles highlight the previously \emph{invisible} regions of occluded objects. }
	\label{fig:results_invisible}
\end{figure*} 

\begin{table}[tb!]
	\centering
	\renewcommand{\arraystretch}{1.1}
	\setlength\tabcolsep{3pt}
	\caption{\textbf{Object Completion}. $\text{F}_{gt}$ = full ground-truth mask, $\text{V}_{gt}$ = visible ground-truth mask.  For methods provided with $\text{F}_{gt}$, we only evaluate the completion networks. }
	\label{table:completion}
	\begin{tabular}{@{}l |c c c | c |c c c@{}}
		\hlineB{3}
		& \multicolumn{3}{c|}{C1-$\text{F}_{gt}$} & & \multicolumn{3}{c}{C2} \\
		\cline{2-4}\cline{6-8}
		& RMSE & SSIM & PSNR & & RMSE & SSIM & PSNR \\
		\hlineB{3}
		SeGAN & 0.1246 & 0.8140 & 21.42 & \multirow{2}{*}{C2a-$\text{V}_{gt}$} & 0.2390 & 0.6045 & 16.03 \\
		PCNet & 0.1129 & 0.8267 & 23.16 & & 0.2483 & 0.5931 & 15.54 \\
		\cline{1-8}
		DNT & 0.1548 & 0.7642 & 20.32 & & 0.2519 & 0.5721 & 15.10 \\
		PICNet & 0.0927 & 0.8355 & 28.81 & C2b & 0.1401 & 0.7730 &  24.71 \\
		\textbf{CSDNet} & \bf{0.0614} & \bf{0.9179} & \bf{35.24} & & \bf{0.0914} & \bf{0.8768} & \bf{30.45} \\
		\hlineB{2.5}
	\end{tabular}	
\end{table}

As can be seen in Table~\ref{table:syn_depth}, all instantiations of our model outperformed baselines as well as previous models. Unlike SeGAN~\citep{ehsani2018segan} and PCNet \citep{zhan2020self}, our final model explicitly predicts the occlusion labels of instances, which improved the OAP substantially. While MLC \citep{qi2019amodal} predicts the instance occlusion order in a network, it only contains one layer for binary occlusion / non-occlusion labeling. In contrast, our method provides a fully structural decomposition of a scene in multiple steps. Additionally, we observed that our model achieves better performance with a higher IoU threshold for selecting the segmented mask (closer match to the ground-truth masks). We further observed that the occlusion relationships of small objects are difficult to infer in our method. However, the \emph{Y-axis} ordering method had similar performance under various metrics as it only depends on the locations of objects. Note that our depth ordering does \emph{not} rely on any ground-truth that is used in~\citep{ehsani2018segan,zhan2020self}.

\begin{table*}[tb!]
	\caption{\textbf{Ablations} for joint optimization. In each table, we fixed one model for one subtask and trained different models for the other subtask. Better performance in one task can improve the performance in the other, which demonstrates the joint training of two tasks with layer-by-layer decomposition contributes to each other.}
	\label{table:ablation}
	\begin{minipage}[t!]{0.48\textwidth}
		(a) \textbf{Effect of different completion methods on instance segmentation (HTC-based decomposition)}. ``sep'' = separate training of the 2 networks, ``w/o'' = without any completion, and ``end'' = joint training.
	\end{minipage}
	\begin{minipage}[t!]{0.48\textwidth}
		(b) \textbf{Effect of different decomposition methods on scene completion (Patch-Attention PICNet)}. Better scene decomposition improved scene completion, likewise with joint training being most effective.
	\end{minipage}
	\begin{minipage}[t!]{0.50\textwidth}
		\begin{center}
			\renewcommand{\arraystretch}{1.1}
			\setlength\tabcolsep{5pt}
			\scalebox{1.0}{\begin{tabular}{@{}l | c |c c c| c c c@{}}
					\hlineB{3}
					& train & AP & AP$_{50}$ & AP$_{75}$ & AP$_{S}$ & AP$_M$ & AP$_L$ \\
					\hlineB{3}
					gt & - & 56.0$^*$ & 67.9$^*$ & 59.3$^*$ & 19.6$^*$ & 53.4$^*$ & 59.5$^*$ \\
					w/o & - &  36.8 & 52.6 & 38.3 & 10.8 & 31.6 & 38.2\\
					\hline
					PICNet-point & sep & 40.8 & 63.0 & 43.5 & 12.6 & 37.2 & 43.7 \\
					PICNet-patch & sep & 43.8 & 60.7 & 46.6 & 11.0 & 36.3 & 45.5\\
					\hline
					PICNet-point & end & 47.7 & 63.2 & 50.6 & 14.9 & 41.7  & 51.3 \\
					PICNet-patch & end & \bf{50.3} & \bf{67.7} & \bf{53.4} & \bf{17.4} & \bf{44.2} & \bf{53.1}\\
					\hlineB{2.5}
			\end{tabular}}
		\end{center}
	\end{minipage}	
	\hfill
	\begin{minipage}[t!]{0.50\textwidth}
		\begin{center}
			%\scriptsize
			\renewcommand{\arraystretch}{1.2}
			\setlength\tabcolsep{5pt}
			\scalebox{1.0}{\begin{tabular}{@{}l|c|c c c@{}}
					\hlineB{3}
					& train & RMSE & SSIM & PSNR \\
					\hlineB{3}
					gt & - & 0.0614$^*$ & 0.9179$^*$ & 35.24$^*$\\
					\hline 
					M-RCNN\citep{he2017mask} & sep & 0.1520 & 0.7781 & 22.34 \\
					HTC \citep{chen2019hybrid} & sep & 0.1496 & 0.7637 & 26.75\\
					\hline
					M-RCNN\citep{he2017mask}  & end & 0.1345 & 0.7824& 27.31\\
					HTC \citep{chen2019hybrid} & end & \bf{0.0914} & \bf{0.8768} & \bf{30.45} \\
					\hlineB{2.5}
			\end{tabular}}
		\end{center}
	\end{minipage}
\end{table*}

\par\medskip\noindent\textbf{Object completion.} We finally evaluated the quality of generated appearances. We compared our results to those from SeGAN \citep{ehsani2018segan}, Dhamo \etal \citep{Dhamo2019iccv} (abbrev.\ as DNT), PCNet \citep{zhan2020self} and PICNet \citep{zheng2019pluralistic} (original point-attention) in Table \ref{table:completion}. We evaluated different conditions of: C1) when the ground-truth full mask $\text{F}_{gt}$ is provided to all methods, C2a) when the ground-truth visible mask $\text{V}_{gt}$ is the input to SeGAN and PCNet, and C2b) when an RGB image is the only input to other methods. C2a-$\text{V}_{gt}$ is considered because SeGAN and PCNet assumes that a predefined mask is provided as input. 

In C1-$\text{F}_{gt}$, CSDNet substantially outperformed the other methods. In C2, even when given only RGB images \emph{without} ground-truth masks, our method worked better than SeGAN and PCNet with $\text{V}_{gt}$. One important reason for the strong performance of CSDNet is the \emph{occlusion reasoning} component, which constraints the completed shapes of partly occluded objects based on the global scene context and other objects \emph{during testing}. 

Qualitative results are visualized in Fig.~\ref{fig:results_invisible}. We noted that SeGAN worked well only when ground-truth amodal masks $\text{F}_{gt}$ were available to accurately label which parts were \emph{invisible} that needed filling in, while DNT generated blurry results from simultaneously predicting RGB appearance and depth maps in one network, which is not an ideal approach~\citep{zamir2018taskonomy}. The PCNet \citep{zhan2020self} can not correctly repair the object shape as it trained without ground-truth object shape and appearance. Our CSDNet performed much better on background completion, as it only masked fully visible objects in each step instead of all objects at a go, so that \emph{earlier completed information propagates to later steps}.

\begin{figure*}[hb!]
	\centering
	\includegraphics[width=\linewidth, height=0.16\textheight]{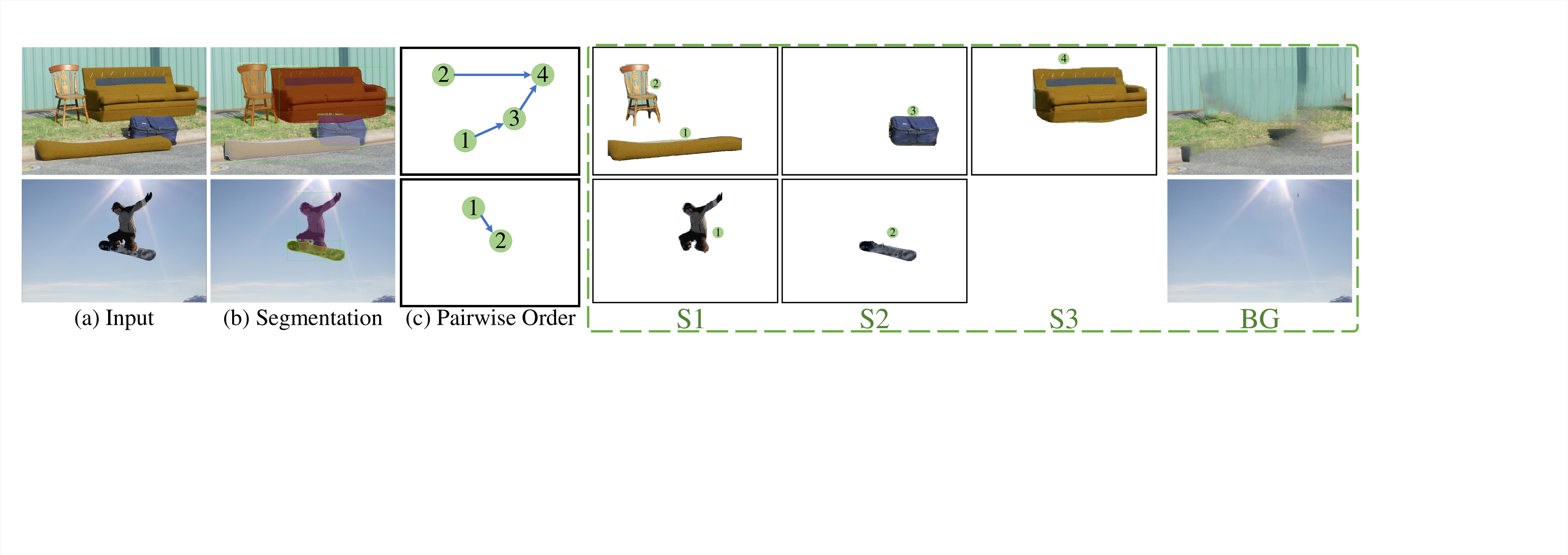}
	\caption{\textbf{Layer-by-layer completed scene decomposition on natural images}. (a) Inputs. (b) Final amodal instance segmentation. (c) Inferred directed graph for pairwise occlusion order. (d) Columns labeled S1-3 show the decomposed instances with completed appearance in each step.}
	\label{fig:results_real_csd}	
\end{figure*}

\subsubsection{Ablation Studies}

To demonstrate the two tasks can contribute to a better scene understanding system by jointly optimizing, instead of solving them isolated, we ran a number of ablations. 

\par\medskip\noindent\textbf{Does better \emph{completion} help decomposition?} We show quantitative results for a fixed decomposition network (layered HTC \citep{chen2019hybrid} with two completion methods in Table~\ref{table:ablation}(a). Without any completion (``w/o''), segmented results were naturally bad (``36.8'' \emph{vs} ``50.3'') as it had to handle empty masked regions. More interestingly, even if advanced methods were used to generate visual completion, the isolated training of the decomposition and completion networks led to degraded performance. This suggests that even when generated imagery looks good visually, there is still a domain or semantic gap to the original visible pixels, and thus flaws and artifacts will affect the next segmentation step. The PICNet with patch attention provides better completed results than the original point attention PICNet \citep{zheng2019pluralistic}, resulting in a large improvement (``50.3'' \emph{vs} ``47.7'') of amodal instance segmentation. 

\par\medskip\noindent\textbf{Does better \emph{decomposition} help completion?} To answer this, we report the results of using different scene segmentation networks with a same completion network (Patch-attention PICNet~\citep{zheng2019pluralistic}) in Table~\ref{table:ablation}(b). We also first considered the ideal situation that ground-truth segmentation masks were provided in each decomposition step. As shown in Table~\ref{table:ablation}(b), the completion quality significantly improved (RMSE: ``0.0614'', SSIM: ``0.9179'' and PSNR: ``35.24'') as occluded parts were correctly pointed out and the completion network precisely knows which parts need to be completed. HTC~\citep{chen2019hybrid} provided better instance masks than Mask-RCNN~\citep{he2017mask}, which resulted in more accurately completed scene imagery. The best results were with end-to-end jointly training.

\begin{figure*}[tb!]
	\centering
	\setlength{\abovecaptionskip}{-0.cm}
	\setlength{\belowcaptionskip}{-0.cm}
	\includegraphics[width=\linewidth]{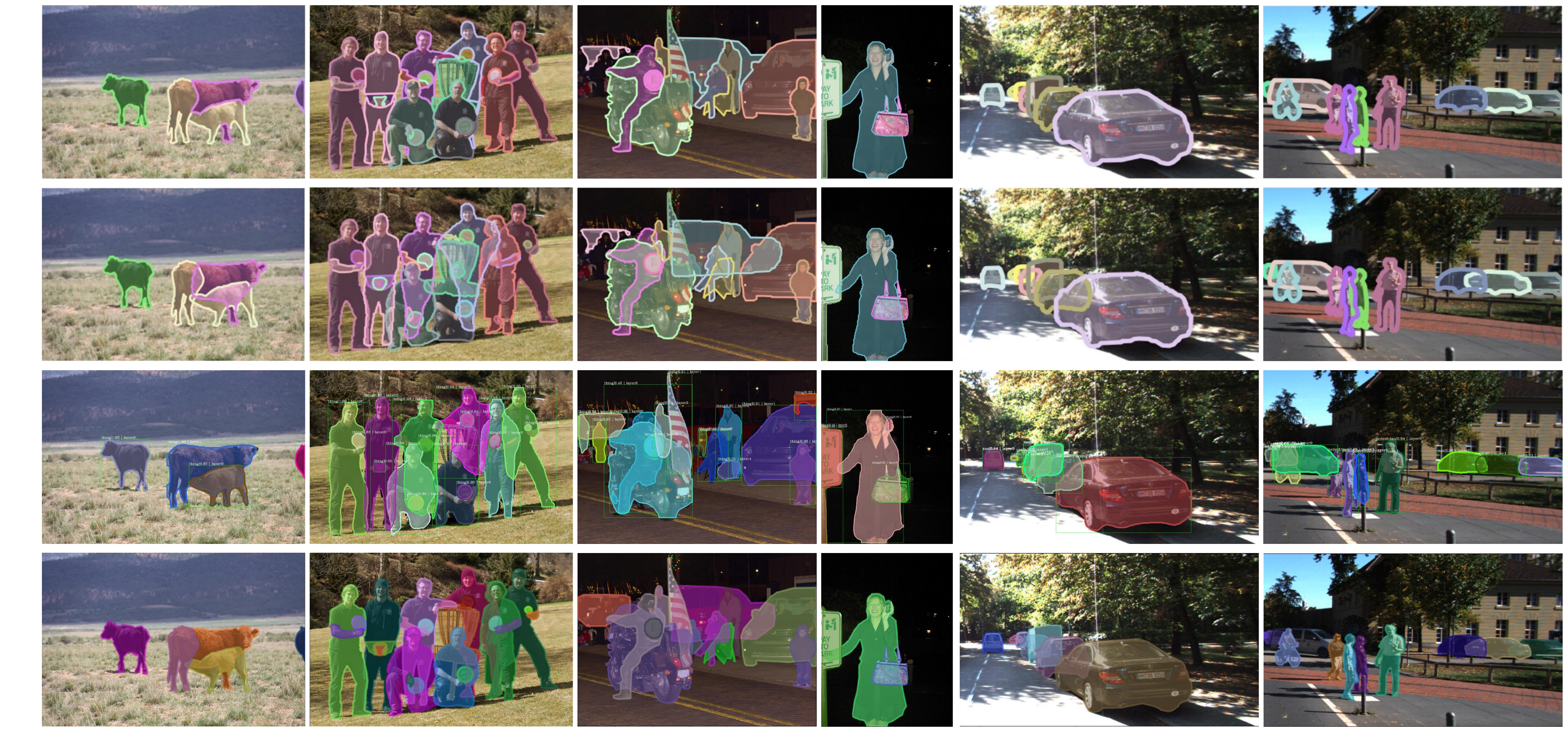}
	\caption{\textbf{Amodal instance segmentation results on natural images}. Our CSDNet learns to predict the intact mask for the occluded objects (\eg animals and human). Note that, unlike PCNet~\citep{zhan2020self}, our model does \emph{not} depend on the visible mask (first row) as input. Hence it can handle some objects without ground-truth annotation, such as two `humans' in the third column and the `smartphone' in the fourth column. }
	\begin{picture}(0,0)
	\put(-243,222){\rotatebox{90}{\footnotesize GT Inmodal}}
	\put(-243,175){\rotatebox{90}{\footnotesize PCNet}}
	\put(-243,121){\rotatebox{90}{\footnotesize Ours}}
	\put(-243,52){\rotatebox{90}{\footnotesize GT Amodal}}
	\end{picture}
	\vspace{-0.3cm}
	\label{fig:results_real_mask}	
\end{figure*}
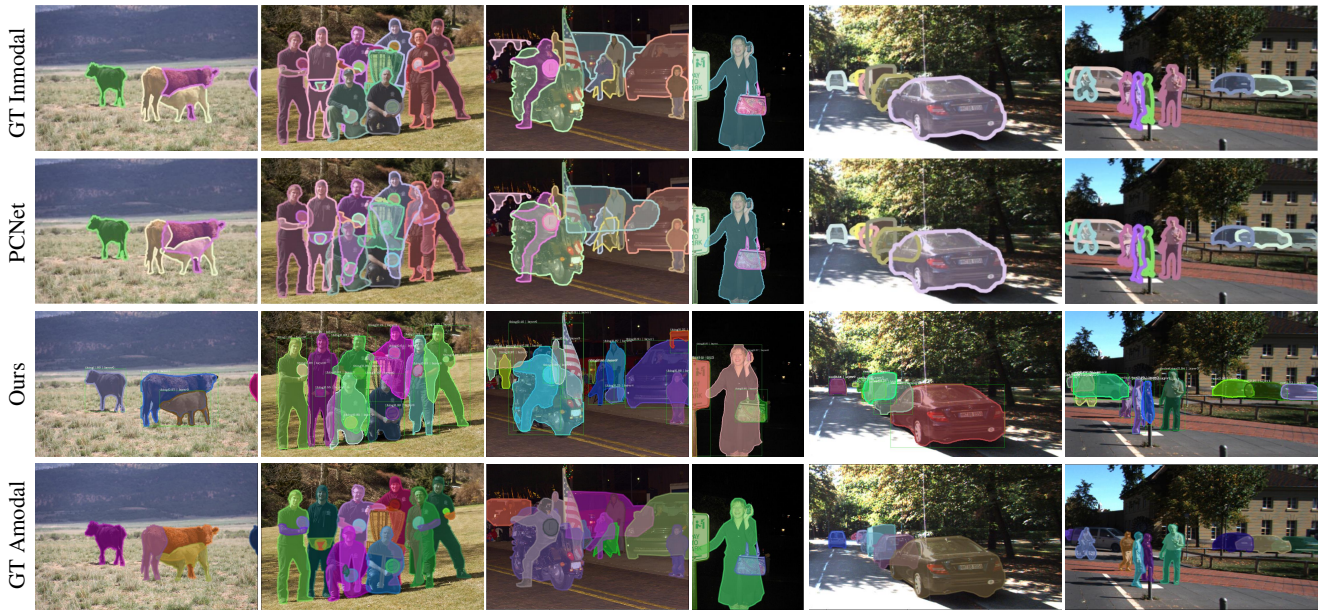

\begin{table*}[tb!]
	\centering
	\caption{\textbf{Amodal Instance Segmentation on COCOA and KINS sets}. The gray color shows results reported in existing works and the others are our reported results by using the released codes and our CSDNet. }
	\label{table:real_seg}
	\renewcommand{\arraystretch}{1.1}
	\begin{tabular}{@{}l|l|l|c|c@{}}
		\hlineB{3.5}
		&  Inputs & SegNet & \makecell[c]{COCOA \\ ($\%$mAP)} & \makecell[c]{KINS\\($\%$mAP)} \\
		\hline 
		Amodel~\citep{zhu2017semantic} & I & Sharp~\citep{pinheiro2016learning} & \cellcolor[rgb]{0.7,0.7,0.7}7.7 & - \\
		Mask-RCNN~\citep{he2017mask} & I & Mask-RCNN~\citep{he2017mask} & \cellcolor[rgb]{0.7,0.7,0.7}31.8 & \cellcolor[rgb]{0.7,0.7,0.7}29.3 \\
		ORCNN~\citep{follmann2019learning} & I & Mask-RCNN~\citep{he2017mask} & \cellcolor[rgb]{0.7,0.7,0.7}33.2 & \cellcolor[rgb]{0.7,0.7,0.7}29.0 \\
		MLC~\citep{qi2019amodal} & I& Mask-RCNN~\citep{he2017mask} & 34.0 & \cellcolor[rgb]{0.7,0.7,0.7}31.1\\
		MLC~\citep{qi2019amodal} & I& HTC~\citep{chen2019hybrid} & 34.4 & 31.6 \\
		%\rowcolor[rgb]{0.6,0.8,1.0}modal testing& V$_{gt}$ & & 49.8 & 55.5\\
		%\rowcolor[rgb]{0.6,0.8,1.0}PCNet~\cite{zhan2020self} & I+V$_{gt}$ &  & 62.6$^*$& 78.4 $^*$\\
		PCNet~\citep{zhan2020self} & I+$\hat{\text{V}}_{pred}$ & Mask-RCNN~\citep{he2017mask} & 30.3 & 28.6 \\
		PCNet~\citep{zhan2020self} & I+$\hat{\text{V}}_{pred}$ & HTC~\citep{chen2019hybrid} & 32.6 & 30.1\\
		\hline
		\textbf{CSDNet} & I & Mask-RCNN~\citep{he2017mask} & 34.1 & 31.5 \\
		\textbf{CSDNet} & I & HTC~\citep{chen2019hybrid} & \textbf{34.8} & \textbf{32.2} \\
		\hlineB{2.5}
	\end{tabular}
\end{table*}

\begin{table*}[tb!]
    \centering
	\caption{\textbf{Instance depth ordering on COCOA and KINS sets}. The blue rows show the results that uses ground-truth annotations as inputs. }
	\label{table:real_order}
	\renewcommand{\arraystretch}{1.1}
	\begin{tabular}{@{}l|l|l|c|c@{}}
		\hlineB{3}
		& \makecell[c]{Ordering \\ Inputs}& \makecell[c]{Ordering \\Algorithm}& \makecell[c]{COCOA \\ (OAP) }& \makecell[c]{KINS \\ (OAP)}\\
		\hlineB{3}
		%& $\text{V}_{gt}$ & Area & \cellcolor[rgb]{0.6,0.8,1.0}55.7 & \cellcolor[rgb]{0.6,0.8,1.0}68.1 \\
		%& $\text{V}_{gt}$ & Y-axis & \cellcolor[rgb]{0.6,0.8,1.0}56.2 & \cellcolor[rgb]{0.6,0.8,1.0}77.5 \\
		%\hline
		OrderNet~\citep{zhu2017semantic} & I+$\text{F}_{gt}$ & Network & \cellcolor[rgb]{0.6,0.8,1.0}88.3 & \cellcolor[rgb]{0.6,0.8,1.0}94.1 \\
		PCNet~\citep{zhan2020self}  & $\text{V}_{gt}$ +$\hat{\text{F}}_{pre}$ & IoU Area & \cellcolor[rgb]{0.6,0.8,1.0}84.6 & \cellcolor[rgb]{0.6,0.8,1.0}86.0 \\
		MLC~\citep{qi2019amodal} & $\text{V}_{gt}$ + $\hat{\text{F}}_{pre}$ & IoU Area & \cellcolor[rgb]{0.6,0.8,1.0}80.3 & \cellcolor[rgb]{0.6,0.8,1.0}82.3 \\
		\textbf{CSDNet} & $\text{V}_{gt}$ + $\hat{\text{F}}_{pre}$ & IoU Area & \cellcolor[rgb]{0.6,0.8,1.0}84.7 & \cellcolor[rgb]{0.6,0.8,1.0}86.4 \\
		\hline
		MLC~\citep{qi2019amodal} & $\hat{\text{V}}_{pred}$ + $\hat{\text{F}}_{pre}$ & IoU Area & 74.2 & 80.2 \\
		MLC~\citep{qi2019amodal} & $\hat{\text{F}}_{pre}$ + layer & layer order$^1$ &  66.5 & 71.8 \\
		PCNet~\citep{zhan2020self} & $\hat{\text{V}}_{pred}$ + $\hat{\text{F}}_{pre}$ & IoU Area &  72.4 & 79.8\\
		\hline
		\textbf{CSDNet} & $\hat{\text{V}}_{pred}$ + $\hat{\text{F}}_{pre}$ & IoU Area & 75.4 & 81.6\\
		\textbf{CSDNet} & $\hat{\text{F}}_{pre}$ + layer & layer order & 80.9 & 82.2\\
		\hlineB{2.5}
	\end{tabular}
\end{table*}

\subsection{Results on Real Datasets}

We now assess our model on real images. Since the ground-truth appearances are unavailable, we only provide the visual \emph{scene manipulation} results in Section~\ref{sec:results_application}, instead of quantitative results for \emph{invisible completion}.

\par\medskip\noindent\textbf{Completed scene decomposition.} In Fig.~\ref{fig:results_real_csd}, we visualize the layer-by-layer completed scene decomposition results on real images. Our CSDNet is able to decompose a scene into completed instances with correct ordering. The originally occluded invisible parts of ``suitcase'', for instance, is completed with full shape and realistic appearance. Note that, our system is a fully scene understanding method that only takes an image as input, without requiring the other manual annotations as ~\citep{ehsani2018segan,zhan2020self}.

\par\medskip\noindent\textbf{Amodal instance segmentation.}
Next, we compare with state-of-the-art methods on amodal instance segmentation. Among these, AmodalMask~\citep{zhu2017semantic} and ORCNN \citep{follmann2019learning} were trained for the COCOA dataset, MLC~\citep{qi2019amodal} works for the KINS dataset, and PCNet~\citep{zhan2020self} is focused on amodal completion (mask completion) rather than amodal instance segmentation (requiring precise visible masks). For a fair comparison, when these methods do not provide results on a given dataset, we trained their models using publicly released code. %For COCOA, we only report the results for ``thing'' category (\eg car, person, chair), because the ``stuff'' category (\eg glass, cloud, water) does not have specific shapes.

Table~\ref{table:real_seg} shows that our results (34.8 mAP and 32.2 mAP) are 0.4 points and 0.6 points higher than the recent MLC using the same segmentation structure (HTC) in COCOA and KINS, respectively. PCNet~\citep{zhan2020self} considers amodal perception in two steps and assumes that visible masks are available. We note that their mAP scores were very high when the visible ground-truth masks were provided. This is because all initial masks were matched to the annotations (without detection and segmentation errors for instances, as shown in Fig.~\ref{fig:results_real_mask}). However, when we used a segmentation network to obtain visible masks $\hat{\text{V}}_{pred}$ for PCNet, the amodal instance segmentation results became lower than other methods, suggesting that it is much harder to segment a visible mask and then complete it.

In Fig.~\ref{fig:results_real_mask}, we compare our CSDNet and PCNet~\citep{zhan2020self}. PCNet only completes the given visible annotated objects which had visible masks. In contrast, our CSDNet produces more contiguous amodal instance segmentation maps even for some unlabeled objects, for instance, the two ``humans'' in the third column. %Furthermore, our model can directly create a deep hierarchical representation of a scene, producing a layer order for each instance. 

\par\medskip\noindent\textbf{Instance depth ordering.}
Finally, we report the instance depth ordering results in Table~\ref{table:real_order}. In order to compare with existing work, we considered two settings: ground-truths provided (blue rows in Table~\ref{table:real_order}), and only RGB images given. The OrderNet obtained the best results as the ground-truth full masks $\text{F}_{gt}$ were given. We note that PCNet and our model achieved comparable performance when the visible ground-truth masks were used. Note that, we only used $\text{V}_{gt}$ for depth ordering, while PCNet utilized the visible mask as input for both mask predication and depth ordering. Furthermore, when no ground-truth annotation was provided as input, our model performed better than MLCand PCNet. 

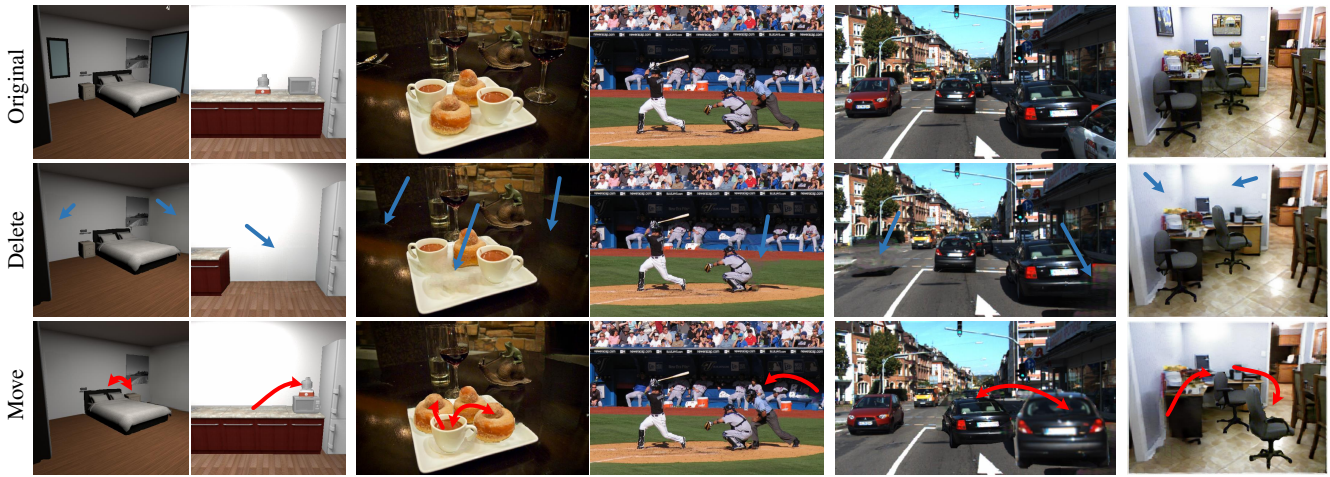
\begin{figure*}[tb!]
	\centering
	\setlength{\abovecaptionskip}{-0.cm}
	\setlength{\belowcaptionskip}{-0.cm}
	\includegraphics[width=\linewidth]{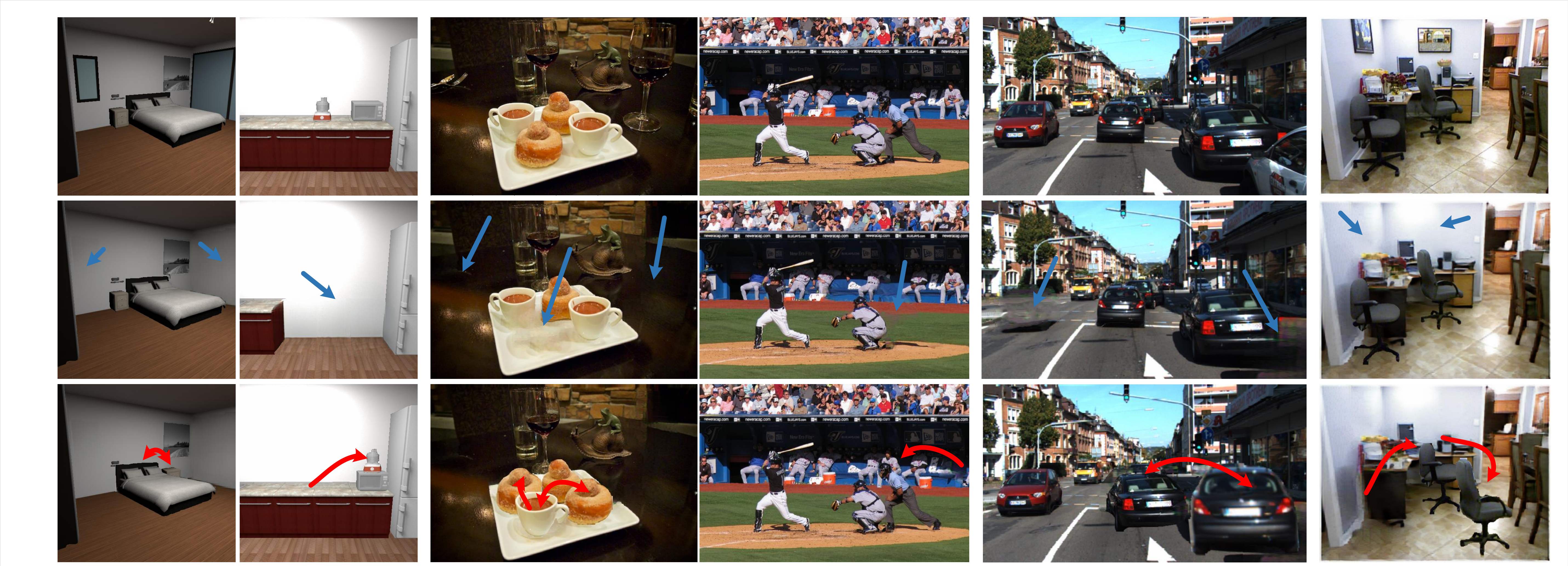}
	\caption{\textbf{Free editing based on the results of our system} on images from various datasets. Note that our method is able to automatically detect, segment and complete the objects in the scene, \emph{without the need for manual interactive masking}, with interactive operations limited to only ``delete'' and ``drag-and-drop''. The blue arrows show object removal, while red arrows show object moving operations. We can observe that the originally \emph{invisible} regions are fully visible after editing.}
	\begin{picture}(0,0)
	\put(-246,186){\rotatebox{90}{\footnotesize Original}}
	\put(-246,131){\rotatebox{90}{\footnotesize Delete}}
	\put(-246,74){\rotatebox{90}{\footnotesize Move}}
	\end{picture}
	\label{fig:results_app}	
\end{figure*}

\subsection{Applications}\label{sec:results_application}
We illustrate some image editing / re-composition applications of this novel task, after we learned to decompose a scene into isolated completed objects together with their spatial occlusion relationships. In Fig.~\ref{fig:results_app}, we visualize some recomposed scenes on various datasets, including our CSD, real COCOA \citep{zhu2017semantic}, KITTI \citep{qi2019amodal} and NYU-v2 \citep{Silberman:ECCV12}. 

In these cases, we directly modified the positions and occlusion ordering of individual objects. For instance, in the first bedroom example, we \emph{deleted} the ``window'', and \emph{moved} the ``bed'' and the ``counter'', which resulted in also \emph{modifying} their occlusion order. Note that all original \emph{invisible} regions were filled in with reasonable appearance. We also tested our model on real NYU-v2~\citep{Silberman:ECCV12} images which do \emph{not} belong to any of the training sets used. As shown in the last column of Fig.~\ref{fig:results_app}, our model was able to detect and segment the object and complete the scene. The ``picture'', for instance, is deleted and filled in with background appearance.

\section{Conclusions}\label{sec:conclusions}

Building on previous inmodal and amodal instance perception work, we explored a higher-level structure scene understanding task that aims to decompose a scene into semantic instances, with completed RGB appearance and spatial occlusion relationships. We presented a layer-by-layer CSDNet, an iterative method to address this novel task. The main motivation behind our method is that fully visible objects, at each step, can be relatively easily detected and selected out without concern about occlusion. To do this, we simplified this complex task to two subtasks: instance segmentation and scene completion. We analyzed CSDNet and compared it with recent works on various datasets. Experimental results show that our model can handle an arbitrary number of objects and is able to generate the appearance of occluded parts. Our model outperformed current state-of-the-art methods that address this problem in one pass. The thorough ablation studies on synthetic data demonstrate that the two subtasks can contribute to each other through the layer-by-layer processing.

\begin{acknowledgements}
	This research was supported by the BeingTogether Centre, a collaboration between Nanyang Technological University (NTU) Singapore and University of North Carolina (UNC) at Chapel Hill. The BeingTogether Centre was supported by the National Research Foundation, Prime Minister’s Office, Singapore under its International Research Centres in Singapore Funding Initiative. This research was also conducted in collaboration with Singapore Telecommunications Limited and partially supported by the Singapore Government through the Industry Alignment Fund ‐- Industry Collaboration Projects Grant and the Monash FIT Start-up Grant.
\end{acknowledgements}


\bibliographystyle{spbasic}      % basic style, author-year citations
\bibliography{egbib}   % name your BibTeX data base

\appendix\twocolumn
\renewcommand{\appendixname}{Appendix~\Alph{section}}
\renewcommand{\theequation}{\thesection.\arabic{equation}}
\setcounter{equation}{0}
\renewcommand{\thefigure}{\thesection.\arabic{figure}}
\setcounter{figure}{0}
\renewcommand{\thetable}{\thesection.\arabic{table}}
\setcounter{table}{0}

\newpage

\section{Experimental Details}

\par\medskip\noindent\textbf{Training.} We trained our model on the synthetic data into three phases: 1) the layered scene decomposition network (Fig.~\ref{fig:framework}(b)) is trained with loss ${L}_{decomp}$ for 24 epochs, where at each layer, re-composited layered ground-truths are used as input. 2) Separately, the completion network (Fig.~\ref{fig:framework}(c)) is trained with loss ${L}_{comp}$ for 50 epochs, wherein the ground-truth layer orders and segmented masks are used to designate the \emph{invisible} regions for completion. 3) Both decomposition and completion networks were trained jointly for 12 epochs, \emph{without relying on ground-truths as input} at any layer (Fig.~\ref{fig:framework}(a)). Doing so allows the scene decomposition network to \emph{learn to cope with flaws} (\eg texture artifacts) in the scene completion network, and vice versa. For each scene, the iteration ends when no more objects are detected, or a maximum 10 iterations is reached.

The training on real data only involved phases 1 and 3, as no ground-truth appearances are available for the invisible parts. The layered decomposition network is trained only for one layer (original image) in phase 1 due to \emph{no} re-composed ground-truth images. Since phase 3 does not rely on ground-truths as input, we trained it layer-by-layer on real images by providing the ``pseudo ground truth'' appearances to calculate the reconstruction loss. To reduce the effect of progressively introduced artifacts in image completion, we used bounding boxes detected in the first layer as proposals for remaining decomposition steps. 

\par\medskip\noindent\textbf{Inference.} During testing, fully visible instances were selected out and assigned an absolute layer order corresponding to the step index $s_k$. In each layer, the decomposition network selects the highest scoring 100 detected boxes for mask segmentation and non-occlusion predication. As we observed that higher object classification scores provided more accurately segmented boundaries, we only selected non-occluded objects with high object classification scores and non-occlusion scores (thresholds of $0.5$ for synthetic images and $0.3$ for real images) among these 100 candidates. We further observed that in some cases, we detected multiple objects with high object classification confidences, yet none were classified as fully visible due to low non-occlusion scores, especially in complex scenes with steps larger than 5. We will then choose the instance with the highest non-occlusion score so that \emph{at least one object is selected at each layer}. When no objects are detected, the iteration stops. 

\begin{figure}[htb!]
	\centering
	\includegraphics[width=\linewidth]{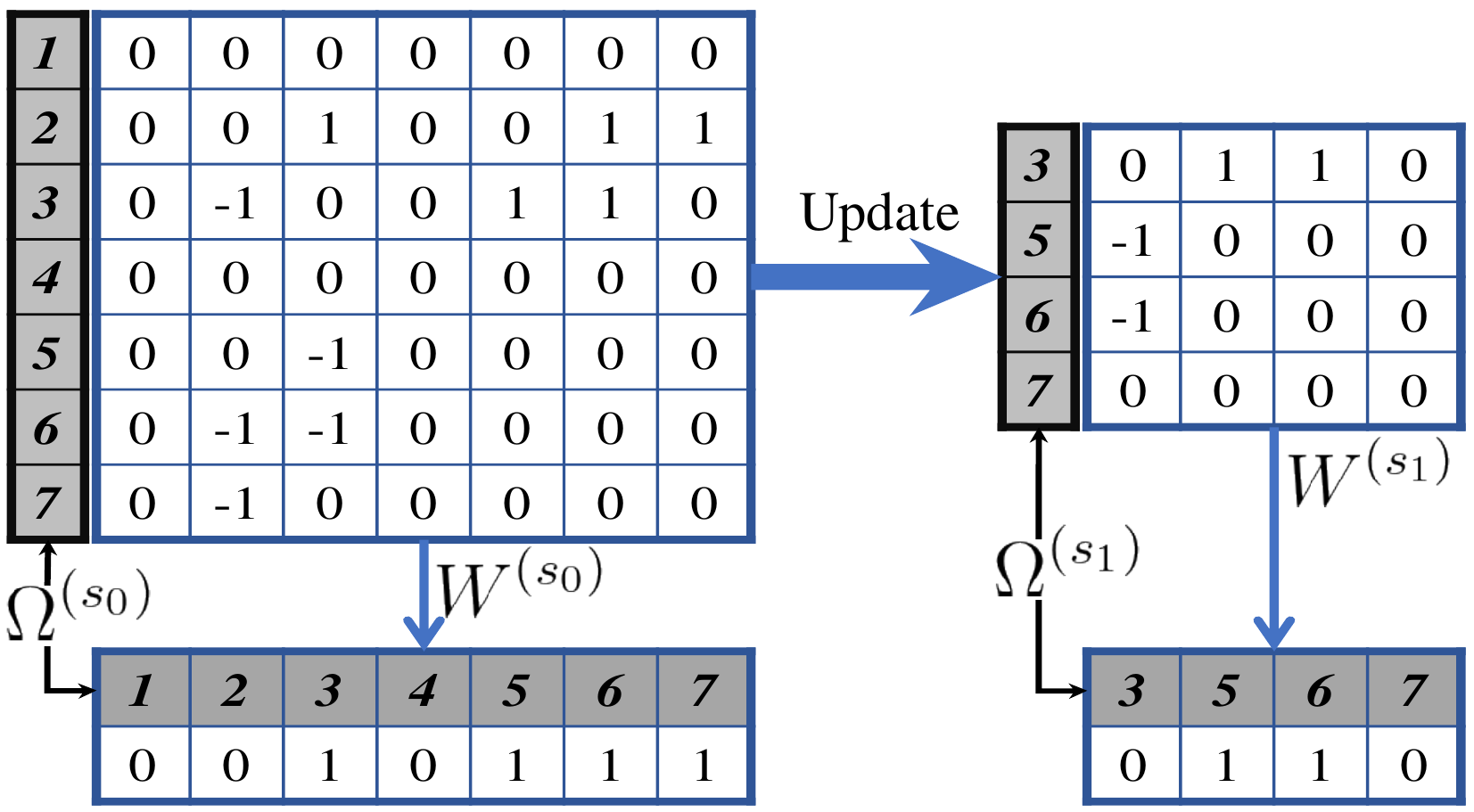}
	\begin{picture}(0,0)
	\put(-96,0){\footnotesize Binary Occlusion Lables}
	\put(36,0){\footnotesize Binary Occlusion Lables}
	\end{picture}
	\caption{\textbf{An illustration of obtaining the ground-truth binary occlusion labels from pairwise order graph $G=(\Omega,W)$ in each step $s_k$.} If the indegree of a vertex is $0$, it will be labeled as $0$, a fully visible instance. Otherwise, the instance will be labeled as $1$, being occluded. When some objects are detected and selected out in the previous step, the object indexes and the corresponding occlusions will be eliminated. }
	\label{fig:layer_update}
\end{figure}
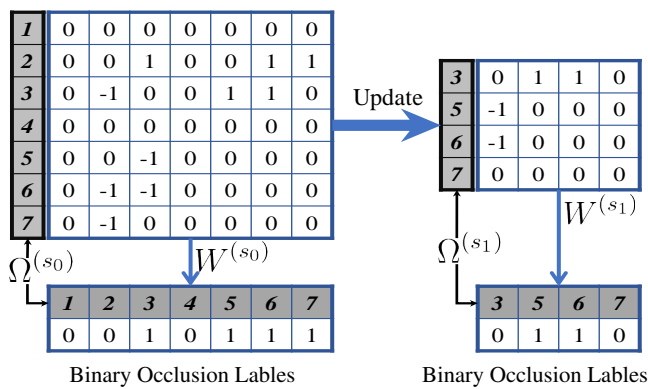 

\begin{figure*}[tb!]
	\centering
	\includegraphics[width=\linewidth]{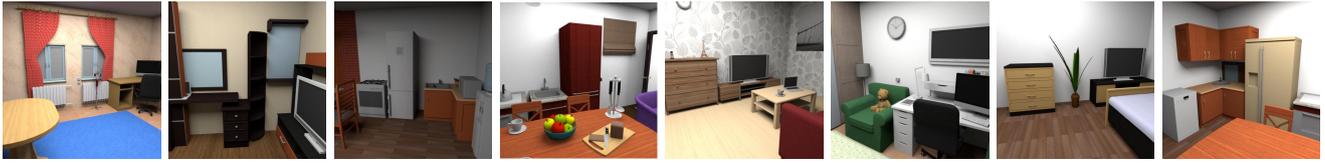}
	\caption{\textbf{Realistic rendered images in the CSD dataset} with various environment and lighting.}
	\label{fig:decom_appendix_dataset_rendering}	
\end{figure*}

\par\medskip\noindent\textbf{Instance depth ordering update.} As illustrated in Fig.~\ref{fig:layer_update}, we calculate the indegree $deg^{-}(\omega)$ (counts of $-1$) of each instance in the matrix. If $deg^{-}(\omega)=0$, meaning no objects are in front of it, its binary occlusion label will be 0. Otherwise, the object is occluded, labeled as 1. At each step, the fully visible objects will be eliminated from the directed graph $G$, and the ground-truth binary occlusion labels will be updated in each step. So if the table (instance \#2) was selected in the previous step, the vertex index $\Omega$ will be updated after the corresponding object $\omega_2$ is deleted from the occlusion matrix. 

\section{Rendering Dataset}\label{appendix_data}

\subsection{Data Rendering}\label{appendix_data_rendering}

Our \textbf{completed scene decomposition (CSD) dataset} was created using Maya \citep{Maya}, based on the SUNCG CAD models \citep{song2017semantic}. The original SUNCG dataset contains 45,622 different houses with realistically modeled rooms. As realistically rendering needs a lot of time (average 1 hour for each house), we only selected 2,456 houses in current work. The set of camera views was based on the original OpenGL-rendering method in SUNCG, but further filtered such that a camera view was only be picked when at least 5 objects appeared in that view. We then realistically rendered RGB images for the selected views. Eight examples are shown in Fig.~\ref{fig:decom_appendix_dataset_rendering} for various room types and lighting environments. Notice that our rendered images are much more realistic than the OpenGL rendered versions from the original SUNCG and likewise in \citep{Dhamo2019iccv}. 

To \emph{visit the invisible}, the supervised method needs ground truth for the original occluded regions of each object. One possible way is to remove the fully visible objects in one layer and re-render the updated scene for the next layer, repeating this for all layers. However, during the training, the fully visible objects are not always correctly detected by the models. Thus, for more robust learning, we need to consider all different combinations of objects and the background. Given $N$ objects, we would need to render $2^N$ images for each view. As we can see from the data statistics presented in Fig. \ref{fig:decom_appendix_dataset_statistic}, an average of 11 objects are visible in each view. Due to slow rendering, we do not have the capacity to render all such scenes (average $2^{11}=2048$ images per view). Instead, we separately rendered each isolated object with the full RGB appearance, as well as the empty room. 

During training, the image of a scene is created by using a combination of the rendered images of these individual objects and the background to create a composed image, based on the remaining objects left after applying the scene decomposition network at each step. Since the room environment is empty for each individual objects during the rendering, the re-composited scenes have lower realism than the original scenes, due to missing shadows and lack of indirect illumination from other objects. In this project, we do not consider the challenges of working with shadows and indirect illumination, leaving those for future research.

\begin{figure*}[h]
	\centering
	\includegraphics[width=\linewidth]{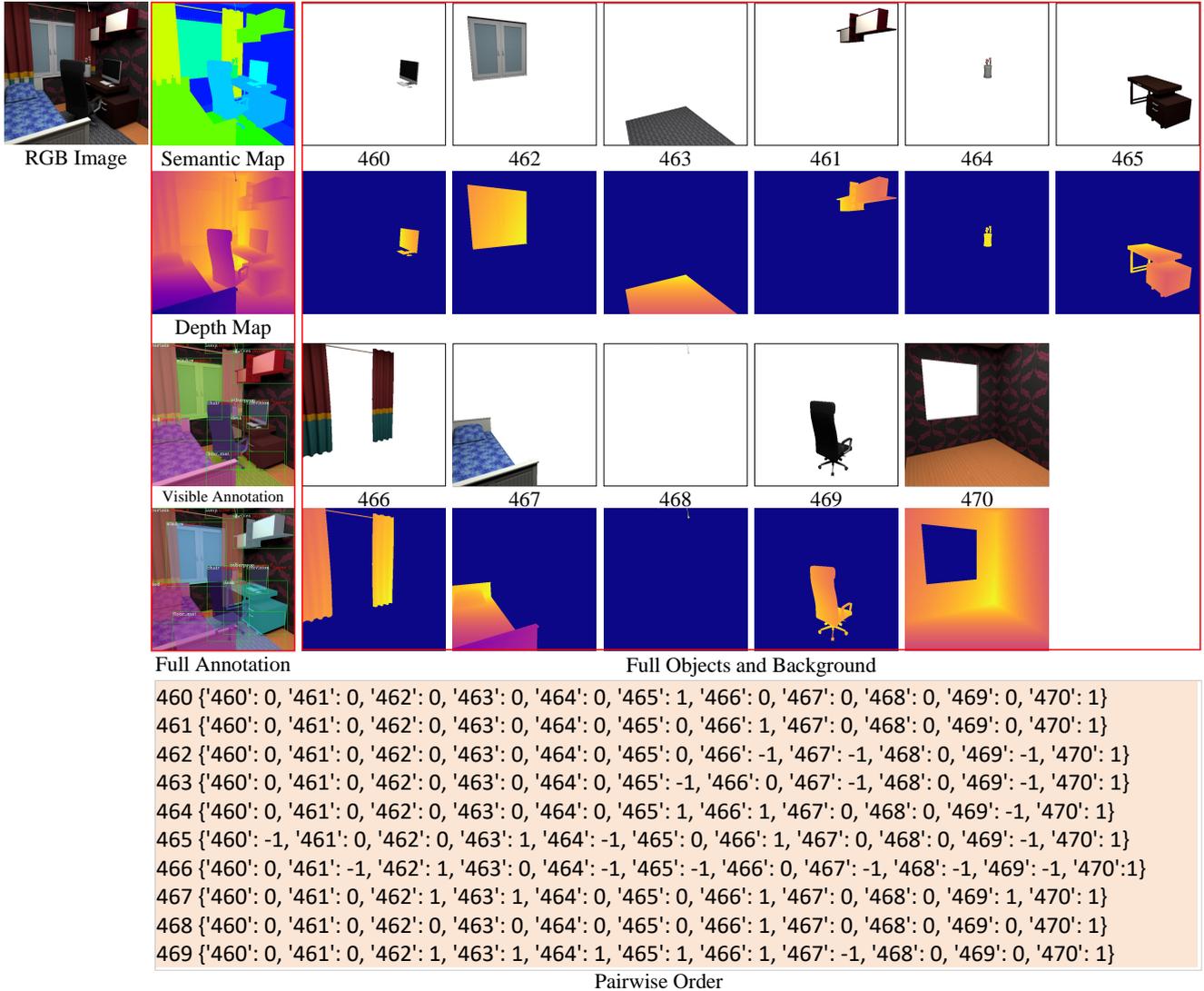}
	\caption{\textbf{Illustration of Data Annotation.} For each rendered image, we have a corresponding semantic map, a depth map, and dense annotation, including class category, bounding box, instance mask, absolute layer order and pairwise order. In addition, for each object, we have a full RGBA image and depth map.}
	\label{fig:decom_appendix_dataset_annotation}
	\vspace{-0.6cm}	
\end{figure*}

\subsection{Data Annotation}\label{appendix_data_annotation}

In Fig.~\ref{fig:decom_appendix_dataset_annotation}, we show one example of a rendered image with rich annotations, consisting of a semantic map, a depth map, visible annotations and full (amodal) annotations. For the semantic maps, we transferred the SUNCG class categories to NYUD-V2 40 categories so that this rendered dataset can be tested on real world images. The depth map is stored in 16-bit format, with the largest indoor depth value at 20m. The class category and layer order (both absolute layer order and pairwise occlusion order) are included for visible annotations and full annotations. The visible annotations also contain the visible bounding-box offset and visible binary mask for each instance. Additionally, we also have the full (amodal) bounding-box offset and completed mask for each individual object.

\paragraph{\bf Pairwise Occlusion Order.} The pairwise order for each object is a vector storing the occlusion relationship between itself and all other objects. We use three numbers $\{-1, 0, 1\}$ to encode the occlusion relationship between two objects --- -1: occluded, 0 : no relationship, 1: front (\ie occluding). As can be seen in Fig.~\ref{fig:decom_appendix_dataset_annotation}, the computer (object number: \#460) does not overlap the shelves (object number: \#461), so the pairwise order is ``0'', indicating these two objects have no occlusion relationship. The computer is however on top of the desk (object number: \#465), hence the pairwise order for $W_{460, 465}$ is ``1'', and conversely the pairwise order for $W_{465, 460}$ is ``-1'', representing that the desk is occluded by the computer. 

\subsection{Data Statistics}\label{appendix_data_statistics}

\begin{figure*}[h]
	\centering
	\includegraphics[width=\linewidth]{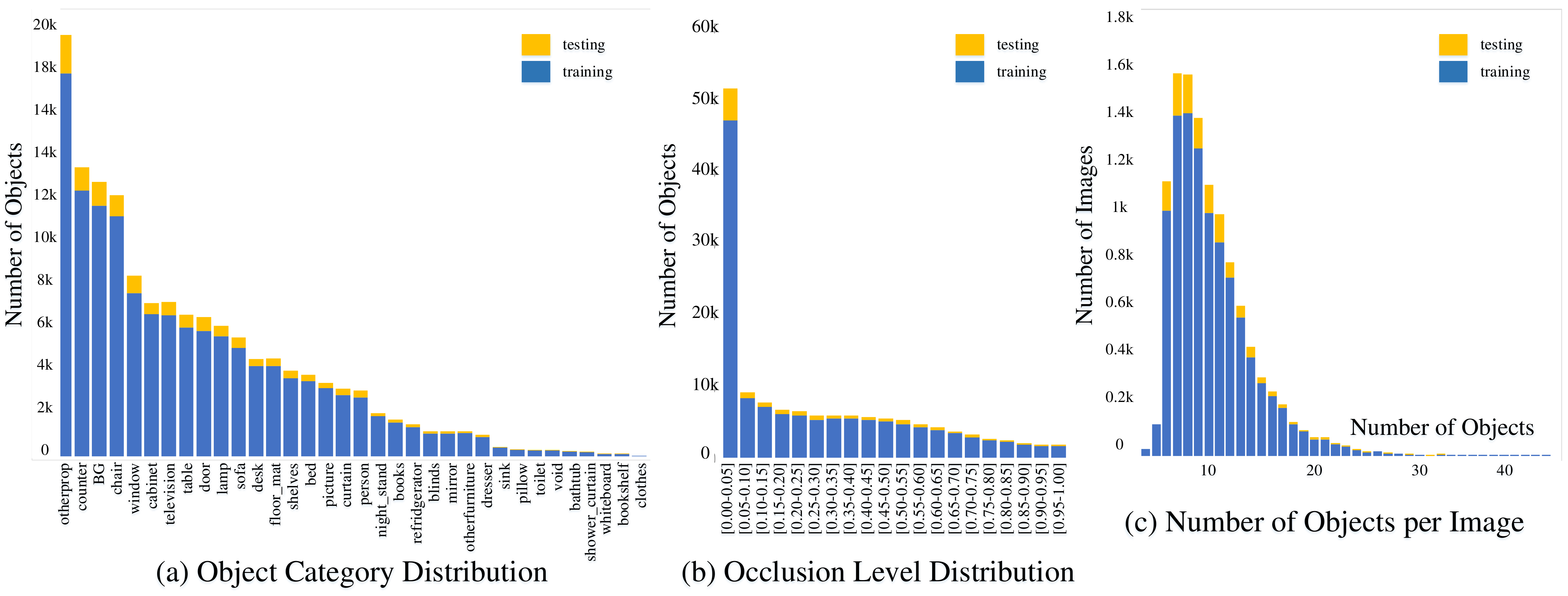}
	\caption{\textbf{Data Statistics.} Left: the object category distribution. Middle: the occlusion level distribution. Right: distribution of number of objects per image. On average there are 11 objects in each room. }
	\label{fig:decom_appendix_dataset_statistic}	
\end{figure*}

In total, there are 11,434 views encompassing 129,336 labeled object instances in our rendered dataset. On average, there are 11 individual objects per view. Among these, 63.58\% objects are partially occluded by other objects and the average occlusion ratio (average IoU between two objects) is 26.27\%.

\paragraph{\bf Object Category Statistics.} Fig.~\ref{fig:decom_appendix_dataset_statistic}(a) shows the overall object category distribution in our CSD dataset. Overall, the distribution is highly similar to the object distribution of NYUD-V2 dataset \cite{Silberman:ECCV12}, containing a diverse set of common furniture and objects in indoor rooms. ``Other props'' and ``Other furniture'' are atypical objects that do not belong in a common category. In particular, "Other props" are small objects that can be easily removed, while "Other furniture" are large objects with more permanent locations. Additionally, we merge floors, ceilings, and walls as ``BG'' in this paper. If the user wants to obtain the separated semantic maps for these structures, these are also available. 

\paragraph{\bf Occlusion Statistics.} The occlusion level is defined as the fraction of overlapping regions between two objects (Intersection over Union, or IOU). We divide the occlusion into 20 levels from highly visible (denoted as [0.00-0.05] fraction of occlusion) to highly invisible (denoted as (0.95-1.00] fraction of occlusion), with 0.05 increment in the fraction of occlusion for each level. Fig.~\ref{fig:decom_appendix_dataset_statistic}(b) shows the occlusion level in our dataset. In general, the distribution of occlusion levels is similar to the distribution in \citep{zhu2017semantic}, where a vast number of the instances are slightly occluded, while only a small number of instances are heavily occluded.

\paragraph{\bf Object Count Distribution.} Fig.~\ref{fig:decom_appendix_dataset_statistic}(c) shows the distribution of the number of objects present per view. On average, there are more than 11 objects in each view. This supports the learning of rich scene contextual information for a completed scene decomposition task, instead of processing each object in isolation. 

\subsection{Data Encoding}\label{appendix_data_encoding}

After we get the views and corresponding dense annotations, we encode the data annotation to COCO format\footnote{http://cocodataset.org}. The annotations are stored using JSON, and the CSD API will be made available for visualizing and utilizing the rendered dataset. The JSON file contains a series of fields, including ``categories'', ``images'' and ``annotations''. One short example is included in the supplementary file named {\bf csd\_short.json}.

\end{document}